\documentclass[letterpaper, 10 pt, conference]{ieeeconf}  %

\IEEEoverridecommandlockouts                              %

\usepackage{lipsum}  

\usepackage{hyperref}

\usepackage{xcolor}

\usepackage{array}  %

\usepackage{graphicx}  %

\usepackage{amsmath}  %
\usepackage{bm}
\usepackage{amsfonts} %
\usepackage{amssymb} %
\usepackage{mathtools}%
\usepackage{glossaries}
\usepackage{siunitx,booktabs}
\usepackage[font=footnotesize,labelfont=bf,tableposition=top]{caption}
\usepackage[export]{adjustbox}

\usepackage{mathtools}

\usepackage{wrapfig}

\usepackage[ruled,vlined,linesnumbered]{algorithm2e}

\SetCommentSty{mycommfont}
\SetKwComment{Comment}{$\triangleright$\ }{}
\let\oldnl\nl%
\newcommand{\nonl}{\renewcommand{\nl}{\let\nl\oldnl}}%

\usepackage[caption=false, font=footnotesize]{subfig}

\newcommand{\citet}{\cite}

\renewcommand{\vec}[1]{\bm{#1}}
\newcommand{\mat}[1]{\mathbf{#1}}

\DeclareMathOperator*{\argmin}{arg\,min}
\DeclareMathOperator*{\argmax}{arg\,max}

\newcommand{\R}{\mathbb{R}}

\newcommand{\zvec}{\vec{z}}

\newcommand{\sethree}{\textsc{SE($3$)}}
\newcommand{\sothree}{\textsc{SO($3$)}}

\newcommand{\grad}{\nabla}

\newcommand{\state}{\vec{s}}
\newcommand{\statedim}{d}

\newcommand{\thetavec}{\vec{\theta}}

\newcommand{\wrt}{w.r.t.}

\DeclarePairedDelimiterX{\infdivx}[2]{(}{)}{%
	#1\;\delimsize\|\;#2%
}

\DeclarePairedDelimiter{\norm}{\lVert}{\rVert}

\newcommand{\E}[2]{\mathbb{E}_{#1}\left[#2\right]}

\newcommand{\transpose}{\intercal}

\newcommand{\muvec}{\vec{\mu}}
\newcommand{\covariance}{\vec{\Sigma}}

\newcommand{\Gaussian}[1]{\mathcal{N}\left(#1\right)}

\newcommand{\horizon}{H}

\newcommand{\task}{\mathcal{O}}

\newcommand{\trajectoryvec}{\vec{\tau}}

\newcommand{\jointpositionvec}{\vec{q}}
\newcommand{\jointvelocityvec}{\dot{\vec{q}}}

\newcommand{\alphacumprod}{\bar{\alpha}}
\newcommand{\betaposterior}{\tilde{\beta}}

\newcommand{\diffusionnoise}{\vec{\varepsilon}}

\newcommand{\data}{\mathcal{D}}
\newcommand{\loss}{\mathcal{L}}

\makeatletter
\newcommand\notsotiny{\@setfontsize\notsotiny\@vipt\@viipt}
\makeatother
\makeatletter
\newcommand{\removelatexerror}{\let\@latex@error\@gobble}
\makeatother

\def\vp{{\bm{p}}}
\def\vq{{\bm{q}}}

\def\vs{{\bm{s}}}

\def\vx{{\bm{x}}}

\def\mK{{\bm{K}}}

\def\mR{{\bm{R}}}

\def\mT{{\bm{T}}}

\def\vtau{{\boldsymbol{\tau}}}

\title{\LARGE \bf
Motion Planning Diffusion:\\
Learning and Planning of Robot Motions with Diffusion Models
}

\author{Jo\~{a}o Carvalho$^{1}$, An T. Le$^{1}$, Mark Baierl$^{1}$, Dorothea Koert$^{1,4}$ and Jan Peters$^{1,2,3,4}$%
\thanks{
This work was funded by the German Federal Ministry of Education and
Research project IKIDA (01IS20045), and
by the German Research Foundation project METRIC4IMITATION (PE 2315/11-1).
$^{1}$Intelligent Autonomous Systems Lab, TU Darmstadt, Germany;
$^{2}$German Research Center for AI (DFKI); 
$^{3}$Hessian.AI;
$^{4}$Centre for Cognitive Science
}
}

\renewcommand{\baselinestretch}{0.98}
\begin{document}

\maketitle
\thispagestyle{empty}
\pagestyle{empty}

\begin{abstract}

Learning priors on trajectory distributions can help accelerate robot motion planning optimization.
Given previously successful plans, learning trajectory generative models as priors for a new planning problem is highly desirable. 
Prior works propose several ways on utilizing this prior to bootstrapping the motion planning problem. Either sampling the prior for initializations or using the prior distribution in a maximum-a-posterior formulation for trajectory optimization. In this work, we propose learning diffusion models as priors. We then can sample directly from the posterior trajectory distribution conditioned on task goals, by leveraging the inverse denoising process of diffusion models. Furthermore, diffusion has been recently shown to effectively encode data multi-modality in high-dimensional settings, which is particularly well-suited for large trajectory dataset. To demonstrate our method efficacy, we compare our proposed method - Motion Planning Diffusion - against several baselines in simulated planar robot and $7$-dof robot arm manipulator environments. To assess the generalization capabilities of our method, we test it in environments with previously unseen obstacles.
Our experiments show that diffusion models are strong priors to encode high-dimensional trajectory distributions of robot motions. 
\href{https://sites.google.com/view/mp-diffusion}{https://sites.google.com/view/mp-diffusion} 

\end{abstract}

\section{Introduction}
Motion planning is a crucial component of autonomous robot systems~\cite{Lav2006planningalgorithms,ratliff2009chomp, Kalakrishnan_RAIIC_2011_stomp, Elbanhawi2014sampling}.
It addresses the problem of finding a feasible, smooth, and collision-free path between a start and a goal point in a robot's configuration space, which can subsequently be executed by a lower-level
controller~\cite{lynch2017modernrobotics}.

Commonly used approaches for motion planning are either sampling \cite{Kavraki1996PRM,Lavalle98rapidly-exploringrandom,kuffner2000rrtconnect, karaman2011sampling} or optimization based~\cite{ratliff2009chomp,Dong2016-gpmp-rss,Mukadam2018-gpmp-ijrr,Kalakrishnan_RAIIC_2011_stomp}. 
Sampling-based methods 
possess a \textit{completeness} property, assuring a global optimum given infinite compute time~\cite{karaman2011sampling}. However, in practice, they often suffer from sample inefficiency and tend to produce non-smooth trajectories~\cite{hauser2010fastsmoothing}.
On the other hand, optimization-based planners optimize initial trajectories via either preconditioned gradient descent~\cite{ratliff2009chomp, Dong2016-gpmp-rss, Mukadam2018-gpmp-ijrr}; or stochastic update rules~\cite{Kalakrishnan_RAIIC_2011_stomp,urain_2022_learning_implicit_priors} and can integrate desired properties such as smoothness as costs to be optimized.
Nevertheless, optimization-based planners depend on a good initialization and can get trapped in local minima due to the non-convexity of complex
problems.
Specifically, they commonly require a good initialization prior and well-tuned hyperparameters to work well~\cite{Mukadam2018-gpmp-ijrr, urain_2022_learning_implicit_priors}.

Recently, learning-based methods have shown promising potential to improve classical motion planning~\cite{wang2021survey}, e.g, by utilizing experience from previously successful 
plans~\cite{ichter2018learning, Qureshi2018motionplanningnet, urain_2022_learning_implicit_priors} or incorporating priors 
from human demonstrations~\cite{koert2016debato,rana2017towardsrobustskill, le2021learning}. 
In particular, sampling from learned prior distributions, conditioned on contexts such as start/goal configurations and environmental variables, can provide good initializations for motion planners~\cite{urain_2022_learning_implicit_priors, urain2022se3diffusion, ortiz2022structureddeep}.

\begin{figure}[t]
    \includegraphics[width=0.237\textwidth]{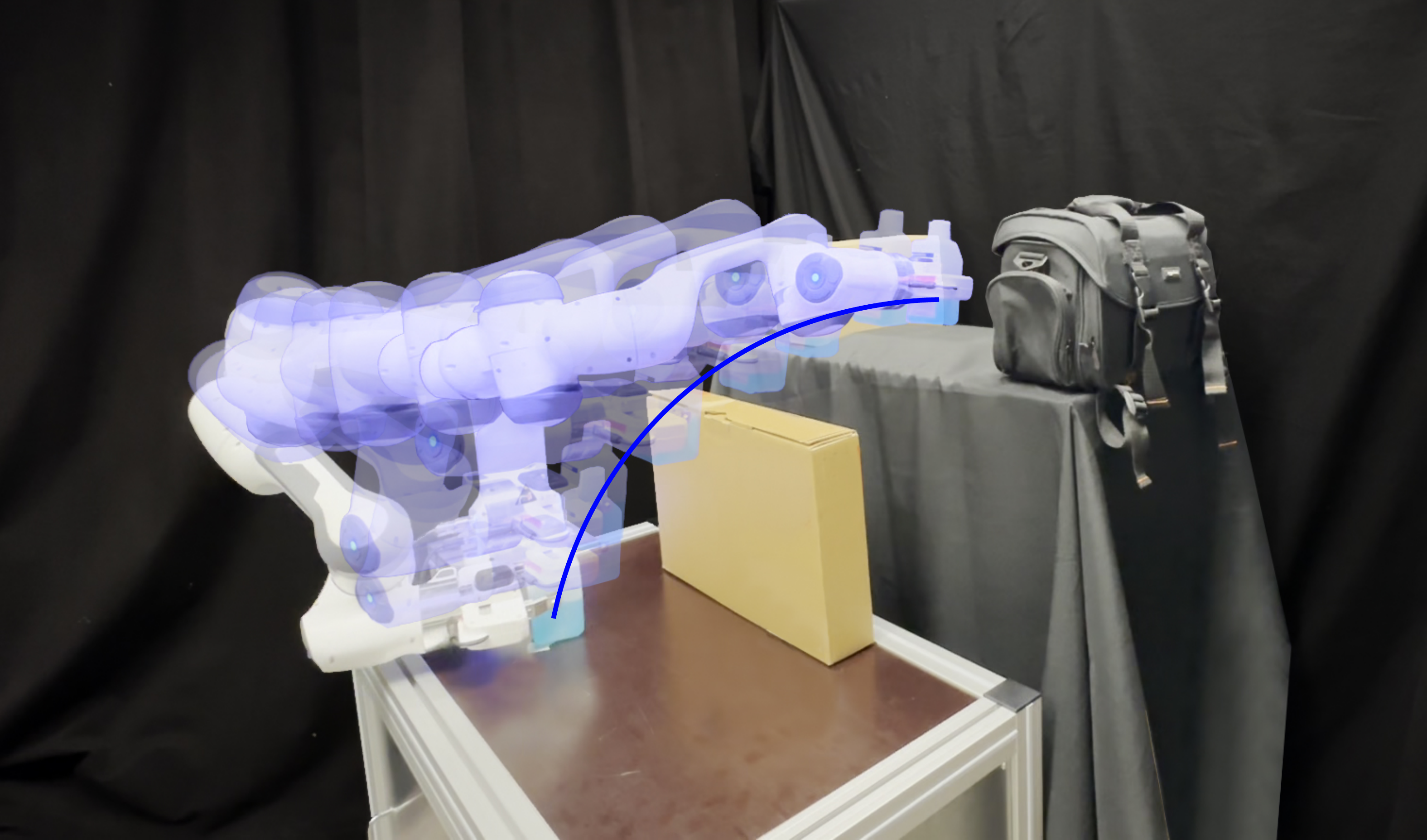}
    \hfill
    \includegraphics[width=0.237\textwidth]{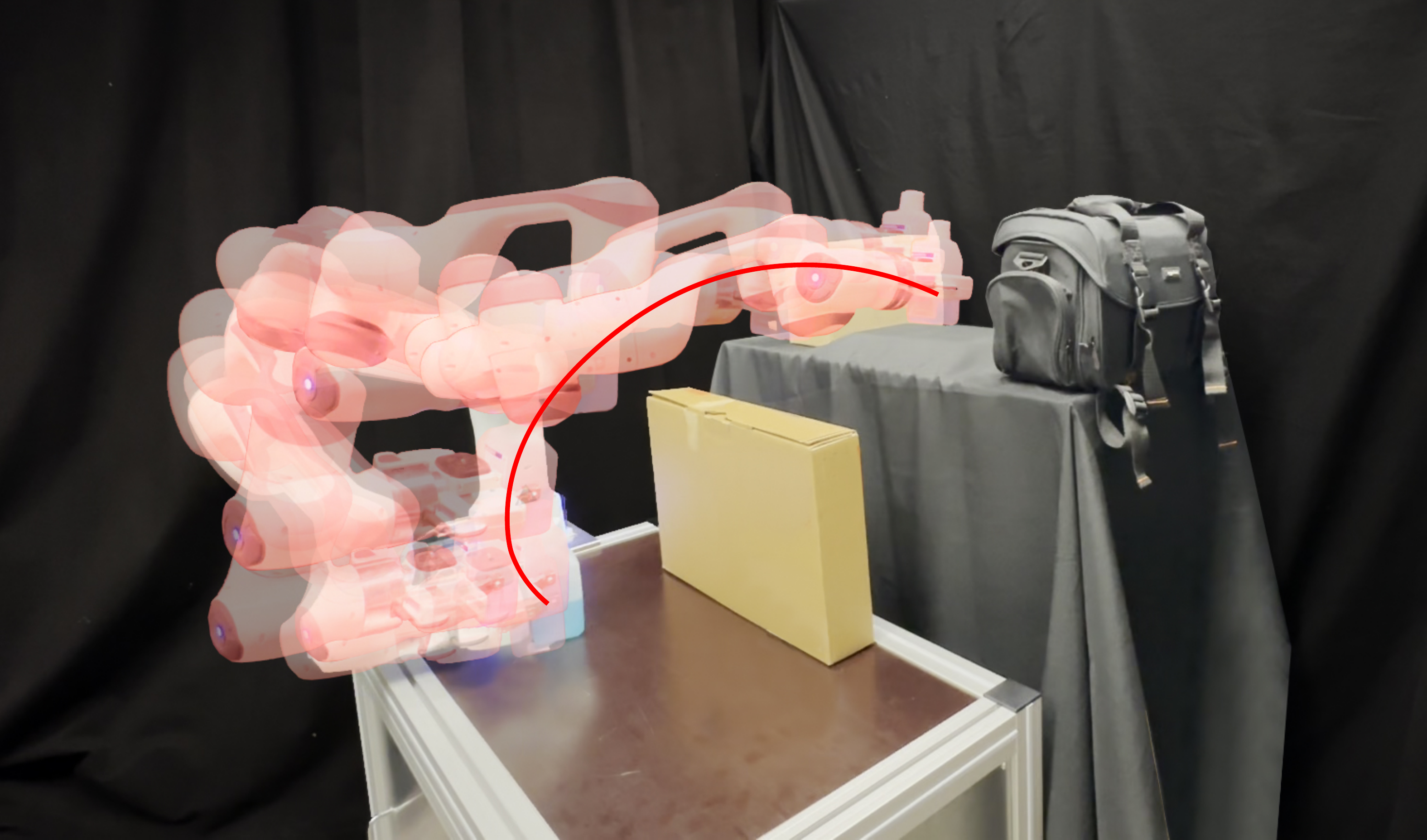}
  \caption{ 
    Execution of motions in the real-world Panda Shelf environment.
    The motion in blue and red start and end at the same configurations, but it is possible to see two modes resulting from sampling trajectories with MPD.
    This environment includes obstacles (represented as boxes in the digital twin), which are not present in the environment used for training (Fig.~\ref{fig:environments:pandapickandplace}).
  }
  \label{fig:real_robot_task}
\vspace{-0.7cm}
\end{figure}

Following the imitation learning perspective~\cite{Takayuki2018imitationlearning}, in this work, we learn the prior model from expert data and incorporate it into optimization-based motion planning. In particular, instead of explicitly sampling from a prior distribution as motion planning initialization, we propose to merge the prior sampling and motion optimization into one algorithm, by leveraging recent formulations in diffusion-based generative models~\cite{Sohl-Dickstein2015diffusion,ho2020denoisingdiffusion}. 
This type of \textit{implicit} model has shown impressive results in modeling multimodal and high-dimensional data, such as image generation~\cite{song2019generativemodeling, ho2020denoisingdiffusion,rombach2021highresolution, Kim_2022_CVPR}
, having superior generative performance (with/out context guidance) compared to previous generative models~\cite{ ramesh2022dalle2,dhariwal2021diffusion}. 
Indeed, these diffusion model properties are particularly well-suited for learning from demonstrations in robotics manipulation settings, where state space dimensions are usually large in manipulators (e.g. full state of position and velocity of Franka Emika Panda arm has $14$ dimensions) and there exist thousands of trajectory samples in the expert dataset. 
Furthermore, as shown later, the diffusion sampling process works well with gradients from standard motion planning costs, allowing for better and more diverse multimodal trajectory solutions.

Our main \textbf{contributions} are:
(1) we learn a trajectory generative model with a diffusion model using expert trajectories generated with an optimal motion planning algorithm;
(2) we formulate the motion planning problem as planning-as-inference by sampling from a posterior distribution leveraging guidance in diffusion models;
(3) to validate our approach we present results in several environments with increasing difficulty;
(4) we empirically demonstrate that learning and sampling from the diffusion model speeds up motion planning without an informed prior and is better than a commonly used generative model.

\vspace{-0.05cm}

\section{Related works} 

There exists a huge body of literature on learning to plan for robotics.
In this section, we focus on discussing related works that combine recent learning methods with classic motion planning approaches (sec. \ref{sec:learning_mp}). Additionally, we provide a short background on applications of diffusion models~\cite{song2019generativemodeling, ho2020denoisingdiffusion} in robotics (sec. \ref{sec:rl_diffusion}).
\vspace{-0.1cm}
\subsection{Learning to plan for motion planning}
\label{sec:learning_mp}

We survey learning methods for both sampling and optimization-based planners. Addressing the sample-inefficiency of sampling-based planners such as Probabilistic Road Maps (PRM)~\cite{Kavraki1996PRM} and Rapid Exploring Random Trees (RRT)~\cite{Lavalle98rapidly-exploringrandom}, several works have proposed learning conditional sampling distributions using the environment and task information as context variables. For instance,~\citet{ichter2018learning} learns to generate collision-free samples using a Conditional Variational AutoEncoder (CVAE) conditioned on an occupancy map. Interestingly,~\citet{johnson2021motionplanningtransformers} proposes Motion Planning Transformers, which determines informative regions for sampling new nodes for RRT-like methods.
Other works~\cite{Qureshi2018motionplanningnet,qureshi2018deepsmp,wang2020neuralrrt}
learn a conditioned one-step neural planner and promote diverse solutions by adding dropout ensuring stochasticity. In contrast to these methods, we learn a trajectory distribution model to easily introduce trajectory time-correlated constraints such as smoothness, which is important in robot motions.

Contrary to sampling-based planners, trajectory optimization methods directly optimize whole trajectories, aiming for smoothness while satisfying other objective constraints. However, they typically heavily depend on initialization and suffer from local minima, e.g. speeding up through obstacles, and thus learning better initial distributions can speed up and improve the success rate of these methods. Several works use Learning from Demonstration (LfD) (i.e., Behavioral Cloning) to encode trajectory priors~\cite{rana2017towardsrobustskill, koert2016debato, le2021learning}.
They fit a Gaussian Mixture Model (GMM) given a set of demonstrations and then use these as priors for motion optimization. However, GMM fitting typically cannot capture well multi-modal trajectories in high-dimensional spaces. 
For better learning the multi-modality of expert trajectories,~\citet{urain_2022_learning_implicit_priors} learned Energy-Based Models (EBMs) that capture expert data objectives. They formulate the trajectory optimization problem via planning-as-inference to incorporate the EBM priors as planning costs. In contrast to incorporating prior as cost, we directly learn a trajectory generator using diffusion models to assist the optimization, where sampling high-dimensional trajectories are difficult for EBMs~\cite{grathwohl2020mcmc}.

\vspace{-0.1cm}
\subsection{Diffusion models for robotics}
\label{sec:rl_diffusion}

Few works have explored using score-based and diffusion models in robotics. At task planning levels,~\citet{Kapelyukh2022-dall-e-bot} recently proposes \textsc{DALL-E-Bot}, which generates a text description from a scene image and then prompts the text-to-image generator DALL-E~\cite{ramesh2022dalle2} for a ``goal scene'', thus identifying the desired poses.~\citet{liu2022structdiffusion} proposes \textit{StructDiffusion} for arranging objects based on language commands, by predicting the object arrangements using a language-conditioned diffusion model. On learning fine-grained trajectory structure, recently, diffusion models have been used to generate human-like motions~\cite{findlay2022ddpmwalking}.
In our previous work~\cite{carvalho2022conditionedsbm}, we showed a simple example of how to use conditioned score-based models to generate trajectories for different environments. Mostly similar to our work,~\citet{urain2022se3diffusion} used diffusion to learn cost functions for jointly optimizing motion and grasping poses in \sethree.
\citet{janner2022diffuser} introduced the Diffuser, a trajectory generative model used for planning in Offline RL and long-horizon tasks.
\citet{ajay2022decision_diffuser} proposes the Decision Diffuser learned over position-only trajectories that can be conditioned with classifier-free guidance skill generation or constraint satisfaction. In contrast to these works, we bring diffusion models closer to motion planners, by incorporating diffusion models as priors combined with differentiable cost likelihoods and also by learning higher trajectory derivatives. We show that, via our formulation, we can directly sample optimal trajectories from the posterior by following the reverse diffusion sampling process, contrasting to sampling from the optimized proposal distribution as in~\cite{urain_2022_learning_implicit_priors}.

\section{Motion Planning Diffusion}

In this section, we describe our method
- Motion Planning Diffusion (MPD).
First, we introduce the motion planning-as-inference perspective, laying the ground for incorporating diffusion into motion planning.
We then explain how to learn trajectory distributions with diffusion models.
Finally, with MPD, we show how to sample from the trajectory posterior via the reverse diffusion process while incorporating the cost likelihood representing other motion planning objectives.
\vspace{-0.1cm}
\subsection{Motion Planning}

Let $\state =[\jointpositionvec^\transpose, \jointvelocityvec^\transpose ]^\transpose \subseteq \R^\statedim$ encode the state of a robot, where $\jointpositionvec$ is the robot's configuration position (e.g. the joint space), $\jointvelocityvec$ is the robot's configuration velocity, $d$ is state space dimension. 
Let $\task$ represent a task (or objective) the robot has to perform in its environment.
A trajectory ${\trajectoryvec \triangleq \left(\state_0, \ldots, \state_{\horizon-1} \right) \in \R^{\horizon \times \statedim}}$ represented as waypoints is a sequence of states in discrete-time with horizon $\horizon$.
In this work, we consider only states and assume a controller brings the robot from state $\state_i$ to $\state_{i+1}$, e.g., using a PD or inverse dynamics controller.

In motion planning, $\task$ typically represents a task that encodes a collision-free path between start and goal states. This objective is commonly represented with a set of costs $c_{i}(\trajectoryvec)$, which encode the robot's start and goal configurations, the trajectory's smoothness, and the collision-free constraint. Optimization-based motion planning formulates the motion planning problem as a trajectory optimization one~\cite{urain_2022_learning_implicit_priors}
\begin{align}
    \trajectoryvec^* = \argmin_{\trajectoryvec} \sum_i \lambda_i c_i(\trajectoryvec),
    \label{eq:trajopt}
\end{align}
where $\lambda_i > 0$ are different weights for the costs. Common approaches for solving this problem are either preconditioned gradient methods~\cite{ratliff2009chomp, Mukadam2018-gpmp-ijrr}, which rely on carefully designed first-order differentiable cost functions; or stochastic methods~\cite{Kalakrishnan_RAIIC_2011_stomp, urain_2022_learning_implicit_priors}, which evaluate the sampled trajectories and update the sampling distribution by weighing the particles on arbitrary non-smooth and non-convex costs.
\vspace{-0.2cm}
\subsection{Motion Planning as Inference}

The connection between trajectory optimization and probabilistic inference is well established~\cite{pmlr-vR4-attias03a, toussaint2009robusttrajopt, levine2018rlcontrolasinference, peters2010reps, watson2019i2c, watson22corl}.
In the planning-as-inference framework, the goal is to sample from the posterior distribution of trajectories given the task objective
\begin{align*}
    p(\trajectoryvec | \task) \propto p(\task | \trajectoryvec) p(\trajectoryvec),
\end{align*}
where $p(\trajectoryvec)$ can be interpreted as a prior over trajectories in the environment, and $p(\task | \trajectoryvec)$ is the likelihood of achieving the task goals. A common assumption is that the likelihood factorizes into independent components~\cite{urain_2022_learning_implicit_priors}
\begin{align}
    p(\task | \trajectoryvec) \propto \prod_{i} p_i(o_i |  \trajectoryvec)^{\lambda_i}
    \label{eq:task_likelihood}
\end{align}
with $\lambda_i > 0$ as temperatures of planning objective distributions~\cite{Le__2022}.
Assuming $p_i$ belongs to the exponential family, we can 
arbitrarily
write $p_i(o_i |  \trajectoryvec) \propto \exp(-c_i(\trajectoryvec))$. Then, performing Maximum-a-Posteriori (MAP) on the trajectory posterior
\begin{align}  \label{eq:mp_objective}
    \trajectoryvec^* &= \argmax_{\trajectoryvec} \log  p(\task | \trajectoryvec) p(\trajectoryvec)  \nonumber \\
    &= \argmax_{\trajectoryvec}  \log \prod_{i} p_i(o_i | \trajectoryvec)^{\lambda_i} + \log  p(\trajectoryvec) \nonumber \\
    &= \argmin_{\trajectoryvec}  \sum_{i} \lambda_i c_i(\trajectoryvec) - \log p(\trajectoryvec)
\end{align}
is equivalent to the motion planning problem~\eqref{eq:trajopt} minus the log of the prior. Contrary to classical optimization-based motion planning, formulating motion planning as an inference problem has several advantages. Notably, the inference framework provides a principal way to introduce informative priors to planning problems, e.g., Gaussian Process Motion Planning (GPMP)~\cite{dong2018sparse, Mukadam2018-gpmp-ijrr} utilizes a Gaussian Process (GP) to encode dynamic feasibility and smoothness as trajectory priors. In this work, we leverage this advantage to incorporate the learned diffusion prior encoded the expert data.

\vspace{-0.1cm}
\subsection{Diffusion Models as Trajectory Generative Models}

Let us consider the unconditional diffusion model on trajectories. Diffusion models transform a trajectory from the data distribution $\trajectoryvec_0 \sim q(\trajectoryvec_0)$ into white Gaussian noise by running a Markovian forward diffusion process ${q(\trajectoryvec_{t} | \trajectoryvec_{t-1}, t) = \Gaussian{\trajectoryvec_{t}; \sqrt{1-\beta_t} \trajectoryvec_{t-1}, \beta_t \mat{I}}}$, where ${t=1, \ldots, N}$ is the diffusion time step (not the trajectory index), $N$ is the number of diffusion steps, and $\beta_t$ is the noise scale at time step $t$.
Common schedules for $\beta$ are linear, cosine or exponential~\cite{ho2020denoisingdiffusion,nichol2021improvedddpm}.
Assuming the data $\trajectoryvec_0$ lives in Euclidean space, the distribution of the diffusion process at time step $t$ is Gaussian and can be written in closed-form as 
$q(\trajectoryvec_{t}|\trajectoryvec_0, t) = \Gaussian{\trajectoryvec_t; \sqrt{\alphacumprod_t}\trajectoryvec_0, (1-\alphacumprod_t) \mat{I}}$, with
$\alpha_t=1-\beta_t$ and $\alphacumprod_t = \prod_{i=1}^t \alpha_i$.
This allows sampling $\trajectoryvec_t$ without running the forward diffusion process~\cite{ho2020denoisingdiffusion}.

The inverse (denoising) process transforms Gaussian noise back to the data distribution through a series of denoising steps $p(\trajectoryvec_{t-1}|\trajectoryvec_{t}, t)$.
Diffusion models approximate this posterior distribution with a parametrized Gaussian ${p_{\theta} (\trajectoryvec_{t-1}|\trajectoryvec_{t}, t) = \Gaussian{\trajectoryvec_{t-1}; \muvec_t = \muvec_{\thetavec}(\trajectoryvec_{t}, t), \covariance_t}}$. 
For simplicity, only the mean of the inverse process is learned, and the covariance is set to $\covariance_t = \sigma_t^2 \mat{I} = \betaposterior_t \mat{I}$, with ${\betaposterior_t = \beta_t (1-\alphacumprod_{t-1})/(1-\alphacumprod_t) }$.
\citet{ho2020denoisingdiffusion} proposed that instead of learning the posterior mean directly, the noise $\diffusionnoise$ can be learned instead, since
${\muvec_{\thetavec}(\trajectoryvec_t, t) = \frac{1}{\sqrt{\alpha_t}} \left( \trajectoryvec_t - \frac{1 - \alpha_t}{\sqrt{1-\alphacumprod_t}} \diffusionnoise_{\thetavec}(\trajectoryvec_t, t) \right)}$,
via a simplified loss function 
\begin{align*}
    \mathcal{L}(\thetavec) = \E{t, \diffusionnoise, \trajectoryvec_0 }{ \| \diffusionnoise - \diffusionnoise_{\thetavec}(\trajectoryvec_t, t) \|_2^2 },
    \label{eq:lossdiffusion}
\end{align*}
with $t\sim \mathcal{U}(1, N)$, $\diffusionnoise \sim \Gaussian{\vec{0}, \mat{I}}$, $\trajectoryvec_0 \sim q(\trajectoryvec_0) $, and ${\trajectoryvec_t = \sqrt{\alphacumprod_t} \trajectoryvec_0 + \sqrt{1-\alphacumprod_t} \diffusionnoise}$.

Following~\cite{janner2022diffuser}, we encode the diffusion model over trajectories with a temporal U-Net, which has proven to be a reasonable architecture for diffusion models over trajectories. 
Please consult~\cite{janner2022diffuser} for details on the network architecture.

\vspace{-0.1cm}
\subsection{Optimal sampling with guidance}
\label{subsec:optimal-sampling-guidance}
We describe how to directly sample from the posterior $p(\trajectoryvec | \task)$ using diffusion models, which is equivalent to sampling from the prior while biasing the trajectories towards the task likelihood. Note that $\trajectoryvec$ results from the last step of the denoising process $\trajectoryvec \equiv \trajectoryvec_0$. By definition of the Markovian reverse diffusion
$
    p(\trajectoryvec_0 | \task) = p(\trajectoryvec_N | \task) \prod_{t=1}^{N} p(\trajectoryvec_{t-1} | \trajectoryvec_t, t, \task),
$
where $p(\trajectoryvec_N | \task)$ is standard Gaussian noise by definition.
Hence, to sample from  $p(\trajectoryvec_0 | \task)$, we iteratively sample from the task-conditioned posterior
\begin{align}
    p(\trajectoryvec_{t-1} | \trajectoryvec_t, t, \task) \propto p(\trajectoryvec_{t-1} | \trajectoryvec_t, t) p(\task | \trajectoryvec_{t-1}),
\end{align}
where $p(\task | \trajectoryvec_{t-1}) = p(\task |  \trajectoryvec_t, \trajectoryvec_{t-1}, t)$, i.e. we drop the conditioning on $\trajectoryvec_{t}$ and $t$ because the task is only conditioned on the current sample $\trajectoryvec_{t-1}$, and $p_{\thetavec}(\trajectoryvec_{t-1} | \trajectoryvec_t, t)$ is modeled with a diffusion model parametrized by $\thetavec$.

To sample from the task-conditioned posterior $p(\trajectoryvec_{t-1} | \trajectoryvec_t, t, \task)$, we will use a similar technique as in classifier guidance, since the posterior cannot be sampled in closed form~\cite{Sohl-Dickstein2015diffusion, Dhariwal2021diffusionbeatsgans}. Considering the learned denoising prior model over trajectories is Gaussian, its logarithm evaluates to
\begin{align} \label{eq:logprior}
    \log p_{\thetavec}(\trajectoryvec_{t-1} | \trajectoryvec_t, t) & = \log \Gaussian{\trajectoryvec_t; \muvec_t = \muvec_{\thetavec}(\trajectoryvec_t, t), \covariance_t} \\
    & \propto -\frac{1}{2} (\trajectoryvec_{t-1} - \muvec_t)^\transpose \covariance_t^{-1} (\trajectoryvec_{t-1} - \muvec_t). \nonumber
\end{align}
By definition of the noise schedule $\beta_t$, as the denoising step approaches zero, so does the noise covariance $\lim_{t \to 0} \| \covariance_t \| = 0$. 
Therefore, $p_{\thetavec}(\trajectoryvec_{t-1} | \trajectoryvec_t, t)$ concentrates all the mass close to the mean $\muvec_t$, and therefore 
the task log-likelihood is approximated with a first-order Taylor expansion around $\muvec_t$
\begin{align}
    \log p(\task | \trajectoryvec_{t-1}) \approx \log p(\task | \trajectoryvec_{t-1} = \muvec_t ) + (\trajectoryvec_{t-1} - \muvec_t) \vec{g},
    \label{eq:logtaskliklelihood}
\end{align}
with $\vec{g} = \grad_{\trajectoryvec_{t-1}} \log p(\task | \trajectoryvec_{t-1}) \rvert_{\trajectoryvec_{t-1} = \muvec_t}$. Combining~\eqref{eq:logprior} and~\eqref{eq:logtaskliklelihood} we obtain 
\begin{align*}
    & \log p(\trajectoryvec_{t-1} | \trajectoryvec_t, t, \task) \\
    & \propto -\frac{1}{2} (\trajectoryvec_{t-1} - \muvec_t)^\transpose \covariance_t^{-1} (\trajectoryvec_{t-1} - \muvec_t) + (\trajectoryvec_{t-1} - \muvec_t) \vec{g} \\
    & \propto -\frac{1}{2} (\trajectoryvec_{t-1} - \muvec_t - \covariance_t \vec{g})^\transpose \covariance_t^{-1} (\trajectoryvec_{t-1} - \muvec_t - \covariance_t \vec{g}) \\
    & = \log p(\vec{z}), \; \text{with } p(\vec{z}) = \Gaussian{\vec{z}; \muvec_t + \covariance_t \vec{g}, \covariance_t}.
\end{align*}
Hence, sampling from the task-conditioned posterior is equivalent to sampling from a Gaussian distribution with mean and covariance as $p(\vec{z})$.
In the motion planning-as-inference case, we have from \eqref{eq:task_likelihood}
\vspace{-0.2cm}
\begin{align*}
    \vec{g} & = \grad_{\trajectoryvec_{t-1}} \log p(\task | \trajectoryvec_{t-1}) \rvert_{\trajectoryvec_{t-1} = \muvec_t} \\
    &= - \sum_{i} \lambda_i \grad_{\trajectoryvec_{t-1}} c_i(\trajectoryvec_{t-1})\rvert_{\trajectoryvec_{t-1} = \muvec_t},
\end{align*}
where the costs are differentiable \wrt~the trajectory, e.g., smoothness or differentiable signed distance functions.

There are direct benefits to this formulation.
While smoothly sampling trajectories from the diffusion prior, we can bias the samples towards regions that increase the task likelihoods, thus resulting in decreasing overall cost objective in~\eqref{eq:mp_objective}. 
In motion planning, these likelihoods can be collision-free regions, goal sets, and joint limits of the configuration space. In practice, to keep the influence of the task likelihood, we typically drop the covariance $\covariance_t$ scaling (cf. line~\ref{alg:diffusionmodelplanningasinference:newtraj} in Algorithm~\ref{alg:diffusionmodelplanningasinference}), since it approaches zero when $t$ approaches $0$. This choice is equivalent to scaling $\lambda_i = \Tilde{\lambda}_i / \sigma_t$ with time-dependent variances $\sigma_t$, where $\Tilde{\lambda}_i$ are the constant hyperparameters, which can be interpreted as increasing likelihood relevancy when $\sigma_t \rightarrow 0$.
Algorithm~\ref{alg:diffusionmodelplanningasinference} depicts the pseudo-code for training and inference of diffusion models for motion planning.
 
\begin{figure}[t]
\removelatexerror
\begin{algorithm}[H]
\caption{Motion Planning Diffusion}
\label{alg:diffusionmodelplanningasinference}
    \DontPrintSemicolon
    {\nonl \textbf{---------TRAINING---------}} \\
    \KwIn{Collision-free trajectories $\data$, Diffusion model $\diffusionnoise_{\thetavec}$, learning rate $\alpha$, noise schedule terms $\alphacumprod_t$}
    \While{training is not finished}{
        \Comment*[l]{sample a batch of trajectories}
        $\trajectoryvec_0 \sim \data $ , $\diffusionnoise \sim \Gaussian{\vec{0}, \mat{I}}, t\sim \mathcal{U}(1, N) $ \\
        \Comment*[l]{compute the diffusion loss function}
        $\trajectoryvec_t = \sqrt{\alphacumprod_t} \trajectoryvec_0 + \sqrt{1-\alphacumprod_t} \diffusionnoise$\\
        $\loss(\thetavec) =  \| \diffusionnoise - \diffusionnoise_{\thetavec}(\trajectoryvec_t, t) \|_2^2 $ \\        
        \Comment*[l]{gradient update}
        $\theta = \theta + \alpha \grad_{\thetavec} \loss(\thetavec) $
    }

    \BlankLine
    {\nonl \textbf{---------INFERENCE---------}} \\
    \KwIn{Pre-trained diffusion model $\diffusionnoise_{\thetavec}$, start and goal states $(\state_{s}, \state_{g})$, motion planning costs $c_i$, temperatures $\lambda_i$, scheduling terms $(\alpha_t, \alphacumprod_t, \sigma_t)$}
    \Comment*[l]{sample a batch of trajectories}
    $\trajectoryvec_{N} \sim \Gaussian{\vec{0}, \mat{I}} $ \\
    \Comment*[l]{hard set start and goal states}
    $\trajectoryvec_{N}[0] = \state_{s}$, $\trajectoryvec_{N}[H-1] = \state_{g}$ \\ \label{alg:diffusionmodelplanningasinference:hardsetstartgoal}
    \For{$t=N,\ldots,1$}{
        \Comment*[l]{compute the diffusion prior mean}
        $\muvec_t = \frac{1}{\sqrt{\alpha_t}} \left( \trajectoryvec_t - \frac{1-\alpha_t}{\sqrt{1-\alphacumprod_t}} \diffusionnoise_{\theta}(\trajectoryvec_t, t) \right) $ \\
        \Comment*[l]{compute weighted gradient of costs}
        $\vec{g} = - \sum_{i} \lambda_i \grad_{\trajectoryvec_{t-1}} c_i(\trajectoryvec_{t-1}=\muvec_{t}) $ \\
        \Comment*[l]{move trajectories to low cost regions}
        $\trajectoryvec_{t-1} = \muvec_t + \vec{g} + \sigma_t \zvec, \quad \zvec \sim \Gaussian{\vec{0}, \mat{I}}$ \\
        \label{alg:diffusionmodelplanningasinference:newtraj}
        $\trajectoryvec_{t-1}[0] = \state_{s}$, $\trajectoryvec_{t-1}[H-1] = \state_{g}$ \label{alg:diffusionmodelplanningasinference:hardsetstartgoal2}
    }
    \KwOut{optimized batch of trajectories $\trajectoryvec_0$}
\end{algorithm}%
\vspace{-0.8cm}
\end{figure}

\vspace{-0.1cm}
\subsection{Motion planning costs}

We briefly describe the standard motion planning costs used in the objective~\eqref{eq:mp_objective}. Note that for cost functions defined on waypoints, the total cost for a trajectory is the sum of waypoint costs, i.e. $c(\trajectoryvec) = \sum_{i} \sum_{j=0}^{H-1} c_i(\trajectoryvec[j])$.

\textbf{Start and Goal States Cost}.
For the diffusion model, the start ($\state_s$) and goal ($\state_g$) state distribution is represented as a Dirac delta, i.e., one factor of \eqref{eq:task_likelihood} for the start state equates to
$p(\trajectoryvec[0] = \state_s | \trajectoryvec) = \delta_{\trajectoryvec[0] = \state_s}(\trajectoryvec) $. In practice, this is implemented by hard setting the initial and final configurations of the trajectory as in lines~\ref{alg:diffusionmodelplanningasinference:hardsetstartgoal} and~\ref{alg:diffusionmodelplanningasinference:hardsetstartgoal2} of Algorithm~\ref{alg:diffusionmodelplanningasinference}~\cite{janner2022diffuser}.
If the generative model does not support Dirac delta distributions, this cost can also be formulated as a quadratic cost, e.g. for the start state $c_{\state_{s}}(\trajectoryvec) = \| \state_{s} - \trajectoryvec[0] \|^2_2$.

\textbf{Collision Cost}. Similar to GPMP~\cite{Mukadam2018-gpmp-ijrr}, we populate $K$ \textit{collision spheres} on the robot body. Given differentiable forward kinematics implemented in PyTorch~\cite{pytorch2019}, with the kinematics Jacobian computed by auto-differentiation, the obstacle cost for any configuration $\vq$ is
$
    {c_{\textrm{obs}}(\vq) = \frac{1}{K}\sum_{j=1}^K c(\vx(\vq, S_j))}
$
with $\vx(\vq, S_j)$ is the forward kinematics position of the $j^{\text{th}}$-collision sphere. For gradient-based motion optimizers, we design the cost using the differentiable signed-distance function $d(\cdot)$ from the sphere center to the closest obstacle surface (minus the sphere radius) in the task space with a $\epsilon > 0$ margin 
$
    {c(\vx) =
    \begin{cases}
        -d(\vx) + \epsilon & \text{if } d(\vx) \leq \epsilon\\
        0 & \text{if } d(\vx) > \epsilon
    \end{cases}}.
$

\textbf{Self-collision Cost}. We group the collision spheres that belong to the same robot links. Then, we compute the pair-wise link sphere distances. The self-collision cost is the average of the computed pair-wise distances.

\textbf{Joint Limits Cost}. We enforce joint limits (and velocity limits) by computing the L$2$ norm joint violations as costs, with a $\epsilon > 0$ margin on each dimension $i$
\begin{equation*}
\small
    c_{\textrm{limits}}(q_i) =
    \begin{cases}
        \norm{q_{\textrm{min}} + \epsilon - q_i}_2^2 & \text{if } q_i < q_{\textrm{min}} + \epsilon\\
        0 & \text{if } q_{\textrm{min}} + \epsilon \leq q_i \leq q_{\textrm{max}} - \epsilon\\
        \norm{q_{\textrm{max}} - \epsilon - q_i}_2^2 & \text{else }
    \end{cases}.
\end{equation*}

\textbf{End-effector Cost}. The distance on $\sethree$ state space defines the end-effector cost. Given two points ${\mT_1 = [\mR_1, \vp_1]}$ and ${\mT_2 =[\mR_2, \vp_2]}$ in $\sethree$, consisting of a translational and rotational part, we choose the following distance on $\sethree$
$
    d_{\sethree}(\mT_1, \mT_2) = \norm{\vp_1 - \vp_2}_2^2 + \norm{\textrm{LogMap}(\mR_1^\intercal\mR_2)},
$
where $\textrm{LogMap}(\cdot)$ is the operator that maps an element of $\sothree$ to its tangent space~\cite{sola2018micro}. Then, given the current end-effector pose $\mT(\vq)$ computed via forward kinematics and the goal pose $\mT_g$, the end-effector cost is 
\vspace{-0.2cm}
\begin{equation}
    c_{ee}(\vq) = d_{\sethree}(\mT(\vq), \mT_g).
    \label{eq:distance_sethree}
\end{equation}

\textbf{Gaussian Process Cost}. The uncontrolled trajectory distribution can be represented as the zero-mean GP prior~\cite{barfoot2014batch, Mukadam2018-gpmp-ijrr} $q_F(\vtau) = \mathcal{N}(\trajectoryvec; \boldsymbol{0}, \mK)$, with a constant time discretization $\Delta t$, and time-correlated covariance matrix ${\mK=[\mK(i, j)]\big|_{ij, 0\leq i,j \leq H-1},\,\mK(i, j) \in \mathbb{R}^{d \times d}}$. 
We can factorize the GP prior by following~\cite{barfoot2014batch}, i.e., 
$$
    q_F(\vtau) \, \propto \,  \exp\big(-\frac{1}{2}\norm{\vtau}^2_{\mK^{-1}}\big) \propto \prod_{t=0}^{H-1} q_t (\vs_t, \vs_{t+1}),
$$
with each binary GP factor defined as
$$
    q_t (\vs_t, \vs_{t+1}) \propto \exp \Big\{ \hspace{-1mm}-\frac{1}{2} \| \mathbf{\Phi}_{t,t+1} \vs_t - \vs_{t+1} \|^2_{\mathbf{Q}_{t,t+1}^{-1}} \Big\},
$$
with $\mathbf{\Phi}_{t,t+1}$ the state transition matrix, and $\mathbf{Q}_{t,t+1}$ the covariance between time steps $t$ and $t+1$. In this work, we assume a holonomic system model and the state transition and covariance matrices are
$$
\small{
\mathbf{\Phi}_{t,t+1} = \begin{bmatrix}
\mathbf{I} & \Delta t\mathbf{I} \\ 
\mathbf{0} & \mathbf{I}
\end{bmatrix}, \quad
\mathbf{Q}_{t,t+1} = \begin{bmatrix}
\frac{1}{3} \Delta t^3 \mathbf{Q}_c &
\frac{1}{2} \Delta t^2 \mathbf{Q}_c \\ 
\frac{1}{2} \Delta t^2 \mathbf{Q}_c &
\Delta t \mathbf{Q}_c
\end{bmatrix}
}
$$
respectively, with $\mathbf{Q}_c$ the power-spectral density matrix. 
Via the planning-as-inference perspective, the GP cost is
\begin{equation*}
    c_{\textrm{GP}}(\vtau) = -\log q_F(\vtau) = \frac{1}{2} \sum_{t=0}^H  \| \mathbf{\Phi}_{t,t+1} \vs_t - \vs_{t+1}  \|^2_{\mathbf{Q}_{t,t+1}^{-1}}.
\end{equation*}
In essence, this cost promotes dynamic feasibility and smoothness. More details on GP priors can be found in~\cite{barfoot2014batch}.

Note that we can introduce further costs, such as manipulability, depending on task requirements.

\section{Experiments}
\label{sec:experiments}

\begin{figure*}[ht!] 
    \centering
  \subfloat[PointMass2D Dense]{%
       \includegraphics[width=0.2\linewidth]{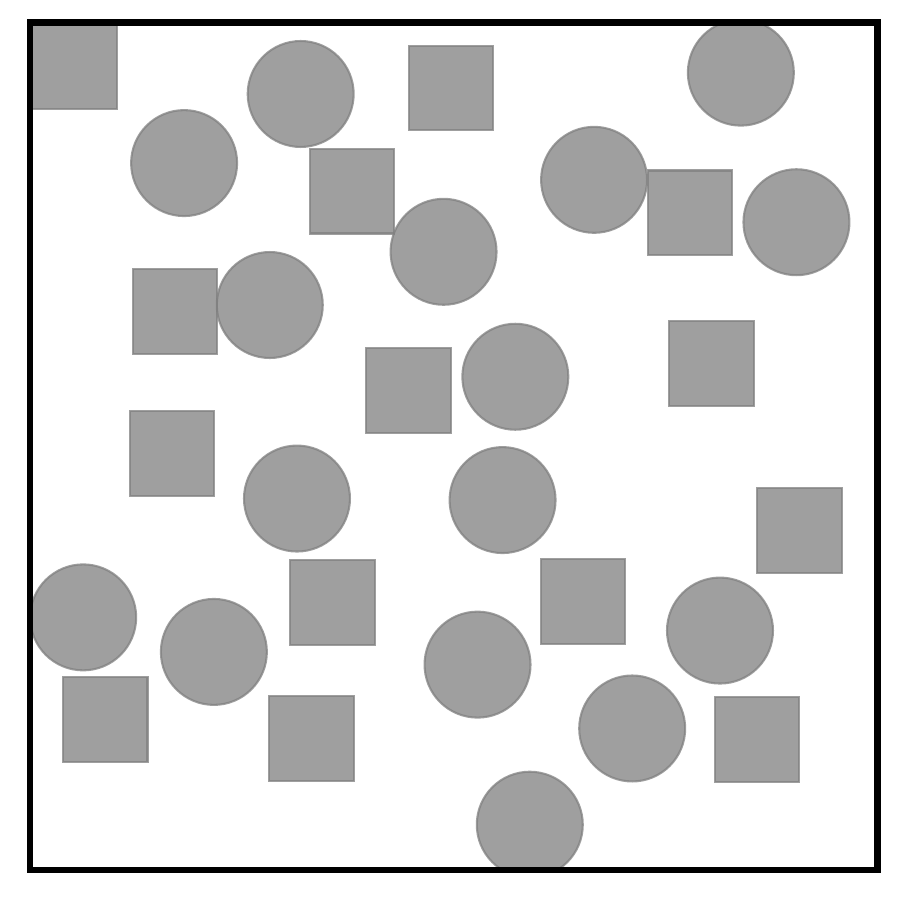}}
       \label{fig:environments:pointmass2d}
    \hfill
  \subfloat[PointMass3D Maze Boxes]{%
        \includegraphics[width=0.2\linewidth]{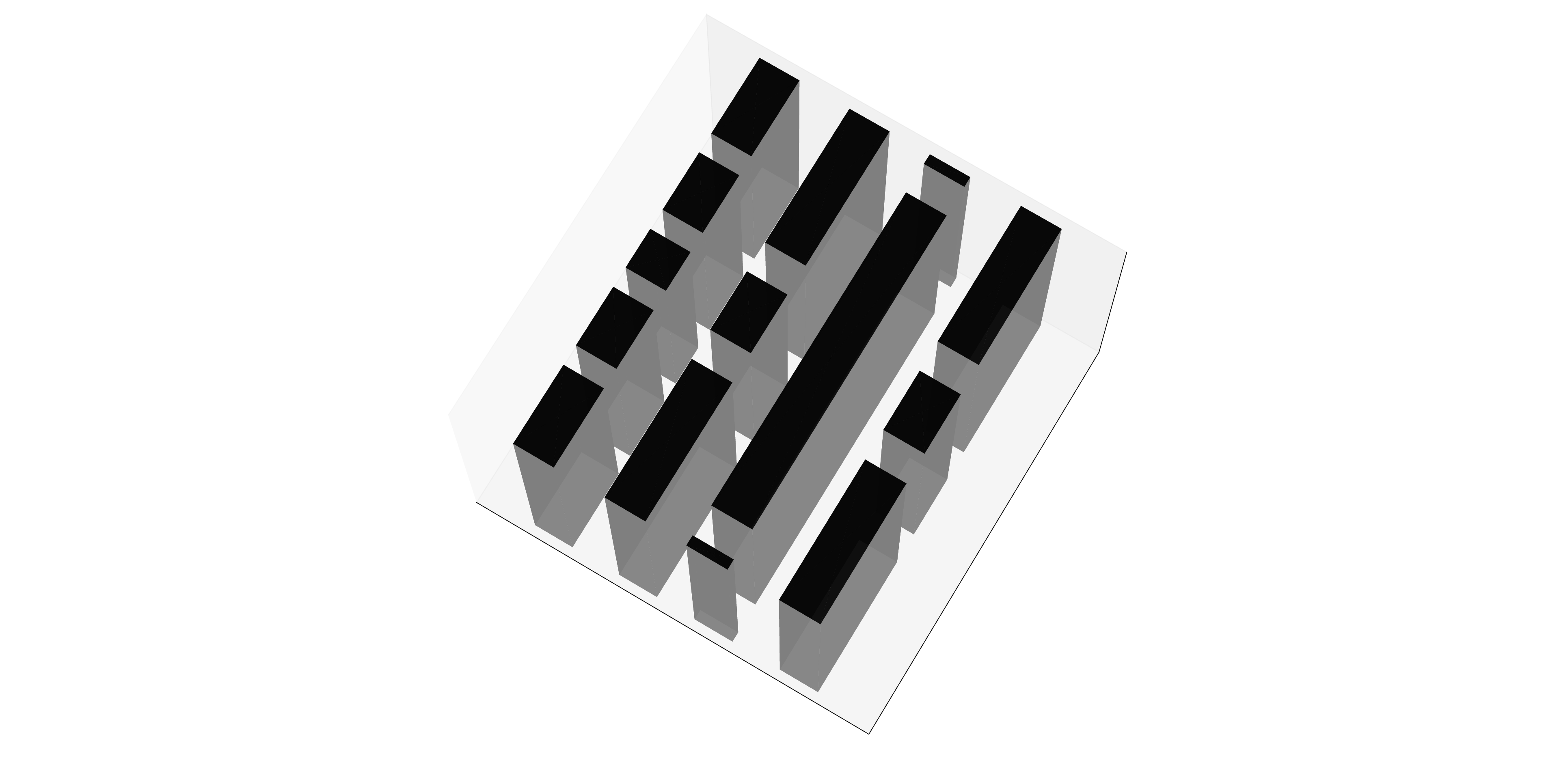}}
        \label{fig:environments:pointmass3d}
    \hfill
  \subfloat[Panda Spheres]{%
        \includegraphics[width=0.2\linewidth]{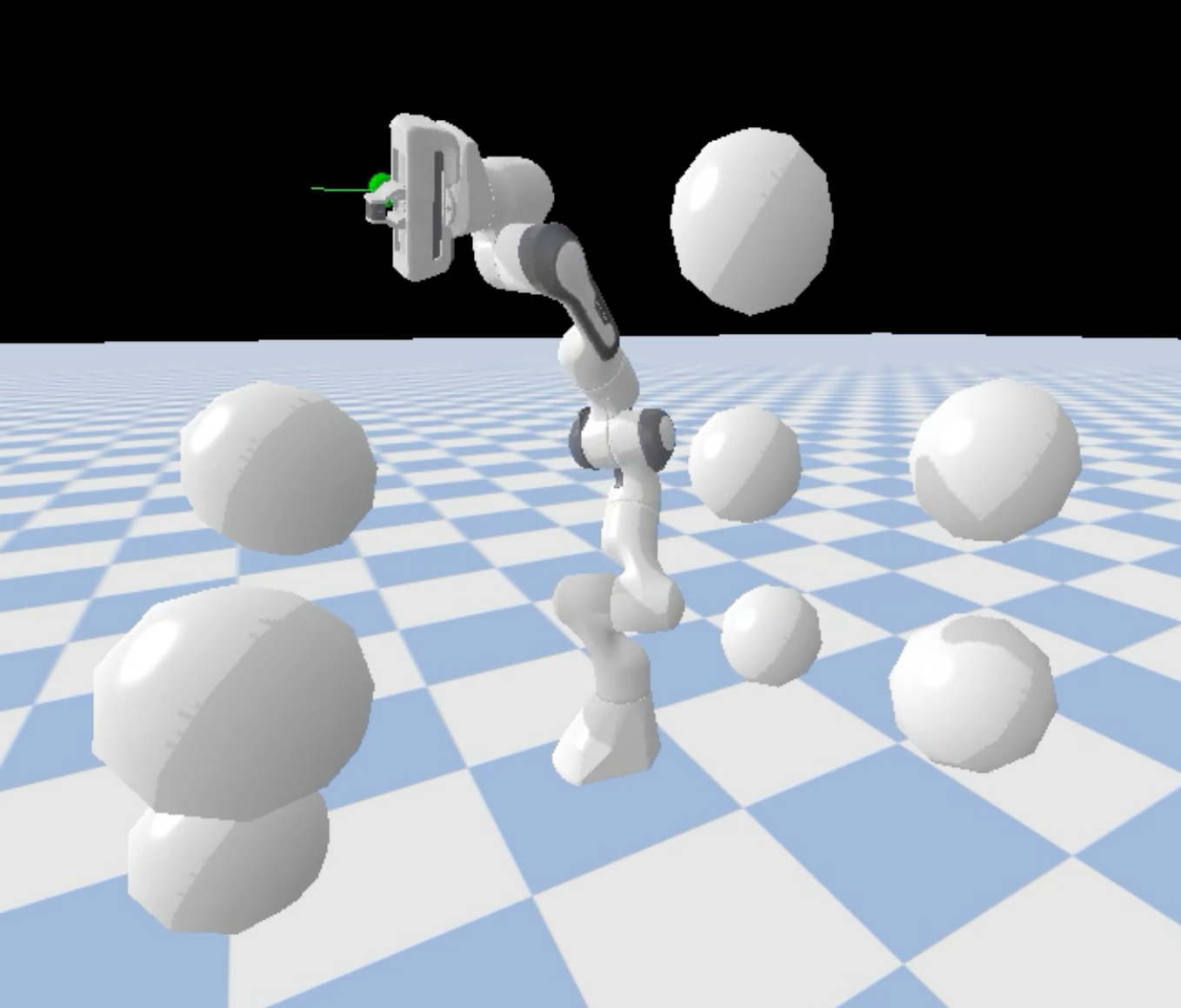}}
        \label{fig:environments:pandaspheres}
    \hfill
  \subfloat[Panda Shelf]{%
        \includegraphics[width=0.2\linewidth]{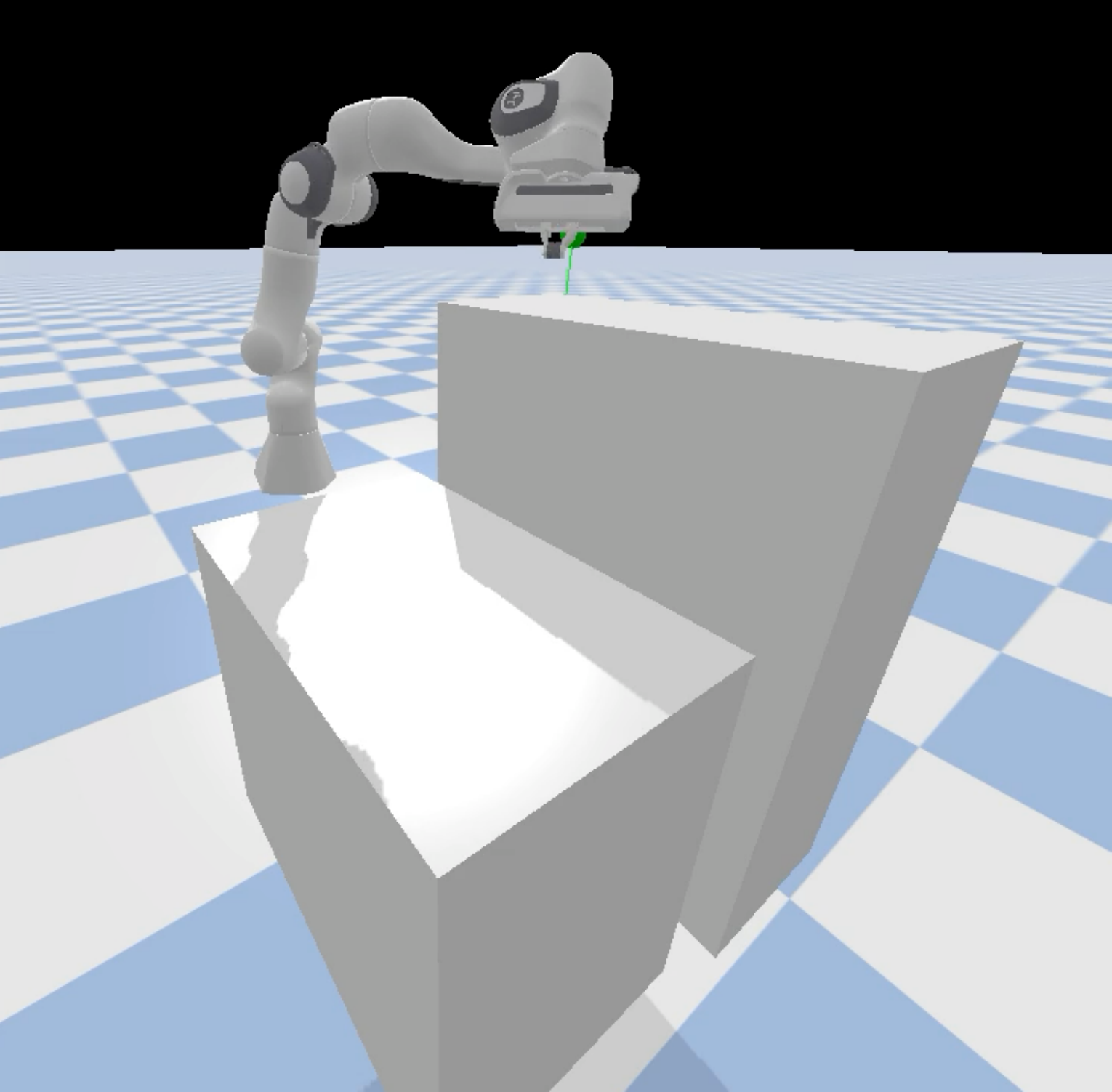}
        \label{fig:environments:pandapickandplace}
        }
  \caption{The environments considered in our experiments include robot navigation tasks of a point mass in $2$D and $3$D, and a $7$-dof Franka Emika Panda manipulator. 
  In (a) and (b), the green and red dots are the initial and goal states, respectively, and the blue line is the result of RRT Connect.
  }
  \label{fig:environments}
  \vspace{-0.3cm}
\end{figure*}

\setlength{\tabcolsep}{4pt}
\begin{table*}[ht!]
\notsotiny

\begin{center}
\captionof{table}{
Motion planning generation benchmarks in the environments of the training set, and additional environments with extra obstacles.
$\uparrow$ means higher is better.
$\downarrow$ means lower is better.
Values highlighted in bold are discussed in the main text.
Legend: RRTC -- RRTConnect, CPrior -- CVAE Prior, DPrior -- Diffusion Prior, CPost -- CVAE Posterior Optimization, MPD -- Motion Planning Diffusion 
}

\vspace{-0.2cm}
\label{tab:benchmark}

\hspace{-0.2cm}\resizebox{\textwidth}{!}{
\begin{tabular*}{\linewidth}{@{}l
@{\extracolsep{\fill}}
r r r r r @{\extracolsep{\fill}} c
r r r r r}
\toprule
\phantom{Var.} &  
\multicolumn{5}{c}{PointMass2D Dense} && \multicolumn{5}{c}{PointMass2D Dense - Extra Obstacles}\\
\cmidrule{2-6}
\cmidrule{8-12} 
& {\textsc{T}$[s] \downarrow$} & {\textsc{S}$[\%] \uparrow$} & {\textsc{I}$[\%] \downarrow$} & {\textsc{PL} $\downarrow$} & {\textsc{Var} $\uparrow$} & {} & {\textsc{T}$[s] \downarrow$} & {\textsc{S}$[\%] \uparrow$} & {\textsc{I}$[\%] \downarrow$} & {\textsc{PL} $\downarrow$} & {\textsc{Var} $\uparrow$} \\
\midrule

RRTC
& $4.7\,\pm 2.5$ & $ 100.0\,\pm .0 $ & $ .0 \,\pm .0 $ & $ {1.9} \,\pm .5 $ & $  $
& 
& $8.0\,\pm 3.5$ & $ 100.0\,\pm .0 $ & $ .0 \,\pm .0 $ & $ {2.2} \,\pm .6 $ & $  $\\

\hline

CPrior
& $.02\,\pm .2$ & $ \bm{46.0}\,\pm 49.5 $ & $ 11.8 \,\pm 12.5 $ & $ 1.5 \,\pm .4 $ & $ \bm{.0} \,\pm .0 $
& 
& $.02\,\pm .2$ & $ \bm{14.0}\,\pm 34.7 $ & $ 24.3 \,\pm 14.7 $ & $ 1.6 \,\pm .4 $ & $ .0 \,\pm .0 $\\

\textbf{DPrior}
& $.3\,\pm .0$ & $ \bm{98.0}\,\pm 14.0 $ & $ 5.5 \,\pm 06.2 $ & $ {1.6} \,\pm .5 $ & $ \bm{1.3} \,\pm 1.0 $
& 
& $.3\,\pm .0$ & $ \bm{60.0}\,\pm 49.0 $ & $ 16.0 \,\pm 10.5 $ & $ {1.5} \,\pm .5 $ & $ 1.3 \,\pm 1.0 $\\

\hline

GPMP
& $26.5\,\pm .2$ & $ \bm{52.0}\,\pm 49.7 $ & $ 1.7 \,\pm 1.3 $ & $ 1.4 \,\pm .4 $ & $ .03 \,\pm .03 $
& 
& $26.5\,\pm .2$ & $ \bm{49.0}\,\pm 50.0 $ & $ 1.1 \,\pm 1.5 $ & $ 1.4 \,\pm .4 $ & $ .03 \,\pm .04 $ \\

RRTC-GPMP
& $38.43\,\pm 30.9$ & $ 100.0\,\pm .0 $ & $ .0 \,\pm 0.0 $ & $ 1.94 \,\pm .5 $ & $ 2.2 \,\pm .9 $
& 
& $42.5\,\pm 7.2$ & $ 100.0\,\pm .0 $ & $ 0.0 \,\pm 0.1 $ & $ 2.0 \,\pm .5 $ & $ 2.7 \,\pm .9 $\\

CPrior-GPMP
& $25.3\,\pm .2$ & $ 71.0\,\pm 45.4 $ & $ 0.9 \,\pm 01.8 $ & $ 1.4 \,\pm .5 $ & $ .01 \,\pm .02 $
& 
& $26.0\,\pm .2$ & $ 53.0\,\pm 49.9 $ & $ 01.1 \,\pm 01.7 $ & $ 1.5 \,\pm .5 $ & $ .02 \,\pm .03 $ \\

\textbf{DPrior-GPMP}
& $26.0\,\pm .3$ & $ \bm{99.0}\,\pm 10.0 $ & $ .3 \,\pm .6 $ & $ 1.6 \,\pm .5 $ & $ 1.4 \,\pm 1.1 $
& 
& $27.0\,\pm .3$ & $ \bm{92.0}\,\pm 27.1 $ & $ .7 \,\pm .8 $ & $ 1.6 \,\pm .5 $ & $ 1.4 \,\pm .9 $\\

\hline

CPost
& $.1\,\pm .1$ & $ 88.0\,\pm 32.5 $ & $ .3 \,\pm .7 $ & $ 1.5 \,\pm .4 $ & $ .01 \,\pm .02 $
& 
& $.1\,\pm .1$ & $ \bm{78.0}\,\pm 41.4 $ & $ .5 \,\pm .7 $ & $ 1.62 \,\pm .5 $ & $ \bm{.01} \,\pm .02 $ \\

\textbf{MPD}
& $.3\,\pm .0$ & $ 99.0\,\pm 10. $ & $ .6 \,\pm 1.2 $ & $ 1.7 \,\pm .5 $ & $ 1.4 \,\pm 1.0 $
& 
& $.3\,\pm .0$ & $ \bm{79.0}\,\pm 40.7 $ & $ 10.3 \,\pm 8.5 $ & $ 1.7 \,\pm .4 $ & $ \bm{1.4} \,\pm 1.0 $ \\

\bottomrule
\end{tabular*}}%
\end{center}

\begin{center}
\hspace{-0.2cm}\resizebox{\textwidth}{!}{
\begin{tabular*}{\linewidth}{@{}l
@{\extracolsep{\fill}}
r r r r r @{\extracolsep{\fill}} c
r r r r r}
\toprule
\phantom{Var.} &  
\multicolumn{5}{c}{PointMass3D Maze Boxes} && \multicolumn{5}{c}{PointMass3D Maze Boxes - Extra Obstacles}\\
\cmidrule{2-6}
\cmidrule{8-12} 
& {\textsc{T}$[s] \downarrow$} & {\textsc{S}$[\%] \uparrow$} & {\textsc{I}$[\%] \downarrow$} & {\textsc{PL} $\downarrow$} & {\textsc{Var} $\uparrow$} & {} & {\textsc{T}$[s] \downarrow$} & {\textsc{S}$[\%] \uparrow$} & {\textsc{I}$[\%] \downarrow$} & {\textsc{PL} $\downarrow$} & {\textsc{Var} $\uparrow$} \\
\midrule

RRTC
& $27.4\,\pm 26.1$ & $ 100.0\,\pm .0 $ & $ .0 \,\pm .0 $ & $ {3.7} \,\pm 1.8 $ & $  $
& 
& $32.0\,\pm 27.4$ & $ 100.0\,\pm .0 $ & $ .0 \,\pm .0 $ & $ {4.0} \,\pm 1.7 $ & $ $\\

\hline

CPrior
& $.02\,\pm .1$ & $ \bm{8.0}\,\pm 27.1 $ & $ 29.1 \,\pm 15.1 $ & $ 2.3 \,\pm .8 $ & $ \bm{.01} \,\pm .02 $
& 
& $.02\,\pm .1$ & $ \bm{4.0}\,\pm 19.6 $ & $ 27.9 \,\pm 14. $ & $ 2.2 \,\pm .8 $ & $ .01 \,\pm .02 $\\

\textbf{DPrior}
& $.3\,\pm .0$ & $ \bm{54.0}\,\pm 49.8 $ & $ 15.5 \,\pm 11.7 $ & $ {2.1} \,\pm .7 $ & $ \bm{4.1} \,\pm 3.9 $
& 
& $.3\,\pm .0$ & $ \bm{51.0}\,\pm 50.0 $ & $ 14.8 \,\pm 8.4 $ & $ {2.2} \,\pm .8 $ & $ 4.1 \,\pm 4.1 $\\

\hline

GPMP
& $47.4\,\pm .1$ & $ \bm{16.0}\,\pm 36.7 $ & $ 32.3 \,\pm 16.6 $ & $ 1.5 \,\pm .5 $ & $ .01 \,\pm .0 $
& 
& $48.5\,\pm .1$ & $ \bm{19.0}\,\pm 39.2 $ & $ 31.2 \,\pm 17.9 $ & $ 1.4 \,\pm .4 $ & $ .01 \,\pm .0 $ \\

RRTC-GPMP
& $114.6\,\pm 56.9$ & $ 100.0\,\pm .0 $ & $ 2.2 \,\pm 1.3 $ & $ 4.1 \,\pm 1.6 $ & $ 5.5 \,\pm 3.0 $
& 
& $111.2\,\pm 52.4$ & $ 100.0\,\pm .0 $ & $ 2.4 \,\pm 1.1 $ & $ 3.9 \,\pm 1.4 $ & $ 5.59 \,\pm 3.1 $\\

CPrior-GPMP
& $47.5\,\pm .3$ & $ 8.0\,\pm 27.1 $ & $ 27.7 \,\pm 15.3 $ & $ 2.2 \,\pm .8 $ & $ .01 \,\pm .03 $
& 
& $47.9\,\pm .3$ & $ 11.0\,\pm 31.3 $ & $ 29.0 \,\pm 16.3 $ & $ 2.1 \,\pm .7 $ & $ .01 \,\pm .03 $ \\

\textbf{DPrior-GPMP}
& $48.3\,\pm .2$ & $ \bm{39.0}\,\pm 48.8 $ & $ 14.4 \,\pm 7.7 $ & $ 2.2 \,\pm .8 $ & $ 4.3 \,\pm 4.1 $
& 
& $49.0\,\pm .2$ & $ \bm{43.0}\,\pm 49.5 $ & $ 15.1 \,\pm 8.5 $ & $ 2.1 \,\pm .8 $ & $ 4.2 \,\pm 3.9 $\\

\hline

CPost
& $.1\,\pm .2$ & $ 50.0\,\pm 50.0 $ & $ 01.4 \,\pm 01.4 $ & $ 2.3 \,\pm .7 $ & $ .02 \,\pm .03 $
& 
& $.1\,\pm .2$ & $ \bm{52.0}\,\pm 50.0 $ & $ 01.3 \,\pm 01.5 $ & $ 2.2 \,\pm .7 $ & $ \bm{.02} \,\pm .03 $ \\

\textbf{MPD}
& $.3\,\pm .01$ & $ 85.0\,\pm 35.7 $ & $ 2.0 \,\pm 1.8 $ & $ 2.4 \,\pm .7 $ & $ 4.2 \,\pm 4.0 $
& 
& $.3\,\pm .01$ & $ \bm{82.0}\,\pm 38.4 $ & $ 3.1 \,\pm 2.1 $ & $ 2.4 \,\pm .8 $ & $ \bm{4.2} \,\pm 4.1 $ \\

\bottomrule
\end{tabular*}}%
\end{center}

\begin{center}
\hspace{-0.2cm}\resizebox{\textwidth}{!}{
\begin{tabular*}{\linewidth}{@{}l
@{\extracolsep{\fill}}
r r r r r @{\extracolsep{\fill}} c
r r r r r}
\toprule
\phantom{Var.} &  
\multicolumn{5}{c}{Panda Spheres} && \multicolumn{5}{c}{Panda Spheres - Extra Obstacles}\\
\cmidrule{2-6}
\cmidrule{8-12} 
& {\textsc{T}$[s] \downarrow$} & {\textsc{S}$[\%] \uparrow$} & {\textsc{I}$[\%] \downarrow$} & {\textsc{PL} $\downarrow$} & {\textsc{Var} $\uparrow$} & {} & {\textsc{T}$[s] \downarrow$} & {\textsc{S}$[\%] \uparrow$} & {\textsc{I}$[\%] \downarrow$} & {\textsc{PL} $\downarrow$} & {\textsc{Var} $\uparrow$} \\
\midrule

RRTC
& $42.9\,\pm 18.8$ & $ 100.0\,\pm .0 $ & $ .04 \,\pm .08 $ & $ {12.1} \,\pm 2.8 $ & $ $
& 
& $90.2\,\pm 103.6$ & $ 100.0\,\pm .0 $ & $ 0.05 \,\pm 0.05 $ & $ {13.2} \,\pm 3.8 $ & $  $\\

\hline

CPrior
& $.02\,\pm .1$ & $ \bm{36.0}\,\pm 48.0 $ & $ 14.9 \,\pm 15.0 $ & $ 5.4 \,\pm 1.0 $ & $ \bm{.02} \,\pm .00 $
& 
& $.02\,\pm .1$ & $ \bm{10.0}\,\pm 30.0 $ & $ 35.2 \,\pm 24.6 $ & $ 5.4 \,\pm 1.2 $ & $ .01 \,\pm .01 $ \\

\textbf{DPrior}
& $.3\,\pm .0$ & $ \bm{88.0}\,\pm 32.5 $ & $ 11.5 \,\pm 6.0 $ & $ {7.8} \,\pm 1.4 $ & $ \bm{17.3} \,\pm 5.0 $
& 
& $.3\,\pm .0$ & $ \bm{78.0}\,\pm 41.4 $ & $ 23.9 \,\pm 7.4 $ & $ {7.5} \,\pm 1.6 $ & $ 18.0 \,\pm 4.2 $ \\

\hline

GPMP
& $194.4\,\pm .1$ & $ \bm{42.0}\,\pm 49.4 $ & $ 4.1 \,\pm 4.8 $ & $ 5.1 \,\pm 1.2 $ & $ .02 \,\pm .05 $
& 
& $194.5\,\pm .2$ & $ \bm{28.0}\,\pm 44.9 $ & $ 9.6 \,\pm 8.2 $ & $ 5.1 \,\pm 1.3 $ & $ .03 \,\pm .12 $ \\

RRTC-GPMP
& $230.4\,\pm 14.4$ & $ 100.0\,\pm .0 $ & $ 02.4 \,\pm 02.7 $ & $ 7.9 \,\pm 1.4 $ & $ 18.7 \,\pm 5.6 $
& 
& $253.6\,\pm 58.7$ & $ 96.0\,\pm 19.6 $ & $ 4.1 \,\pm 2.8 $ & $ 8.0 \,\pm 1.4 $ & $ 20.6 \,\pm 7.2 $\\

CPrior-GPMP
& $191.6\,\pm .3$ & $ 34.0\,\pm 47.4 $ & $ 5.2 \,\pm 5.7 $ & $ 5.1 \,\pm .9 $ & $ .03 \,\pm .06 $
& 
& $194.2\,\pm .1$ & $ 18.0\,\pm 38.4 $ & $ 12.3 \,\pm 10.3 $ & $ 5.1 \,\pm 1.1 $ & $ .04 \,\pm .06 $ \\

\textbf{DPrior-GPMP}
& $192.3\,\pm .1$ & $ \bm{100.0}\,\pm .0 $ & $ 4.0 \,\pm 2.6 $ & $ 6.8 \,\pm 1.2 $ & $ 15.6 \,\pm 3.5 $
& 
& $192.1\,\pm .1$ & $ \bm{82.0}\,\pm 38.4 $ & $ 7.8 \,\pm 3.8 $ & $ 7.1 \,\pm 1.0 $ & $ 16.0 \,\pm 3.6 $ \\

\hline

CPost
& $.8\,\pm .1$ & $ 70.0\,\pm 45.8 $ & $ .7 \,\pm 1.0 $ & $ 8.0 \,\pm 1.3 $ & $ .03 \,\pm .04 $
& 
& $.8\,\pm .1$ & $ \bm{45.0}\,\pm 49.8 $ & $ 1.5 \,\pm 1.7 $ & $ 8.0 \,\pm 1.2 $ & $ \bm{.1} \,\pm .1 $ \\

\textbf{MPD}
& $1.1\,\pm .01$ & $ 100.0\,\pm .0 $ & $ 1.2 \,\pm .8 $ & $ 9.9 \,\pm 1.2 $ & $ 17.4 \,\pm 3.9 $
& 
& $1.1\,\pm .01$ & $ \bm{93.0}\,\pm 25.5 $ & $ 12.5 \,\pm 7.6 $ & $ 10.0 \,\pm 1.2 $ & $ \bm{18.0} \,\pm 4. $ \\

\bottomrule
\end{tabular*}}%
\end{center}

\begin{center}
\hspace{-0.2cm}\resizebox{\textwidth}{!}{
\begin{tabular*}{\linewidth}{@{}l
@{\extracolsep{\fill}}
r r r r r @{\extracolsep{\fill}} c
r r r r r}
\toprule
\phantom{Var.} &  
\multicolumn{5}{c}{Panda Shelf} && \multicolumn{5}{c}{Panda Shelf - Extra Obstacles}\\
\cmidrule{2-6}
\cmidrule{8-12} 
& {\textsc{T}$[s] \downarrow$} & {\textsc{S}$[\%] \uparrow$} & {\textsc{I}$[\%] \downarrow$} & {\textsc{PL} $\downarrow$} & {\textsc{Var} $\uparrow$} & {} & {\textsc{T}$[s] \downarrow$} & {\textsc{S}$[\%] \uparrow$} & {\textsc{I}$[\%] \downarrow$} & {\textsc{PL} $\downarrow$} & {\textsc{Var} $\uparrow$} \\
\midrule

RRTC
& $29.9\,\pm 8.3$ & $ 100.0\,\pm .0 $ & $ 0.04 \,\pm 0.11 $ & $ {11.2} \,\pm 2.4 $ & $  $
& 
& $31.5\,\pm 9.4$ & $ 99.0\,\pm 10.0 $ & $ 2.2 \,\pm 4.5 $ & $ {11.5} \,\pm 2.9 $ & $  $ \\

\hline

CPrior
& $.03\,\pm .2$ & $ \bm{92.}\,\pm 27.1 $ & $ 1.8 \,\pm 6.5 $ & $ 4.8 \,\pm .9 $ & $ \bm{.02} \,\pm .0 $
& 
& $.03\,\pm .1$ & $ {68.0}\,\pm 46.7 $ & $ 8.3 \,\pm 14.1 $ & $ 4.6 \,\pm .8 $ & $ .01 \,\pm .00 $ \\

\textbf{DPrior}
& $.3\,\pm .0$ & $ \bm{100.0}\,\pm .0 $ & $ \bm{3.6} \,\pm 4.8 $ & $ {7.6} \,\pm 1.2 $ & $ \bm{14.5} \,\pm 3.4 $
& 
& $.3\,\pm .0$ & $ \bm{100.0}\,\pm .0 $ & $ \bm{5.9} \,\pm 7.3 $ & $ {7.6} \,\pm 1.3 $ & $ 14.3 \,\pm 3.7 $ \\

\hline

GPMP
& $192.1\,\pm .1$ & $ \bm{88.0}\,\pm 32.5 $ & $ .5 \,\pm 1.6 $ & $ 5.1 \,\pm 1.3 $ & $ .01 \,\pm .02 $
& 
& $193.1\,\pm .13$ & $ \bm{82.}\,\pm 38.42 $ & $ 2.0 \,\pm 5.1 $ & $ 5.1 \,\pm 1.4 $ & $ .01 \,\pm .01 $ \\

RRTC-GPMP
& $218.3\,\pm 7.0$ & $ 100.0\,\pm .0 $ & $ .4 \,\pm .9 $ & $ 8.1 \,\pm 1.1 $ & $ 16.7 \,\pm 3.8 $
& 
& $228.2\,\pm 20.1$ & $ 98.0\,\pm 14.0 $ & $ 1.5 \,\pm 2.1 $ & $ 7.9 \,\pm 1.5 $ & $ 18.3 \,\pm 6.0 $ \\

CPrior-GPMP
& $192.9\,\pm .1$ & $ 94.0\,\pm 23.8 $ & $ .6 \,\pm 2.6 $ & $ 4.4 \,\pm .9 $ & $ .01 \,\pm .00 $
& 
& $194.3\,\pm .2$ & $ 60.\,\pm 49.0 $ & $ 3.3 \,\pm 5.7 $ & $ 4.5 \,\pm .8 $ & $ .03 \,\pm .04 $ \\

\textbf{DPrior-GPMP}
& $192.3\,\pm .1$ & $ \bm{100.0}\,\pm .0 $ & $ 0.9 \,\pm 1.1 $ & $ 7.1 \,\pm 1.1 $ & $ 13.8 \,\pm 2.9 $
& 
& $192.6\,\pm .1$ & $ \bm{98.0}\,\pm 14.0 $ & $ 2.3 \,\pm 2.4 $ & $ 6.9 \,\pm 1.2 $ & $ 14.0 \,\pm 3.1 $ \\

\hline

CPost
& $.8\,\pm .1$ & $ 93.0\,\pm 25.5 $ & $ 0.1 \,\pm 0.4 $ & $ 7.4 \,\pm 1.2 $ & $ .02 \,\pm .04 $
& 
& $.8\,\pm .1$ & $ \bm{83.0}\,\pm 37.6 $ & $ 0.5 \,\pm 1.1 $ & $ 10.8 \,\pm 6.8 $ & $ \bm{2.4} \,\pm 5.2 $ \\

\textbf{MPD}
& $1.0\,\pm .01$ & $ 100.0\,\pm .0 $ & $ 0.6 \,\pm 0.9 $ & $ 9.3 \,\pm 1.0 $ & $ 14.8 \,\pm 3.2 $
& 
& $1.0\,\pm .01$ & $ \bm{99.0}\,\pm 10.0 $ & $ 4.1 \,\pm 4.9 $ & $ 9.4 \,\pm 1.1 $ & $ \bm{15.1} \,\pm 3.5 $ \\

\bottomrule
\end{tabular*}}%
\end{center}

\end{table*}

\begin{figure*}[ht!] 
    \centering
  \subfloat[Diffusion Step 0]{%
       \includegraphics[width=0.191\linewidth]{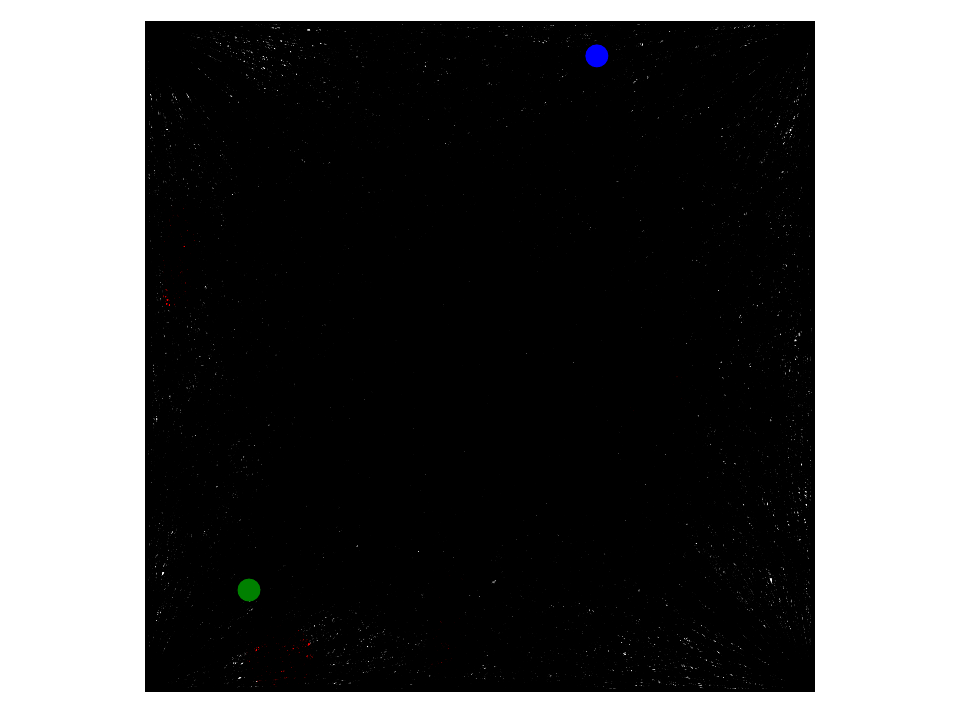}}
       \label{fig:diffusion:0}
    \hfill
  \subfloat[Diffusion Step 10]{%
        \includegraphics[width=0.191\linewidth]{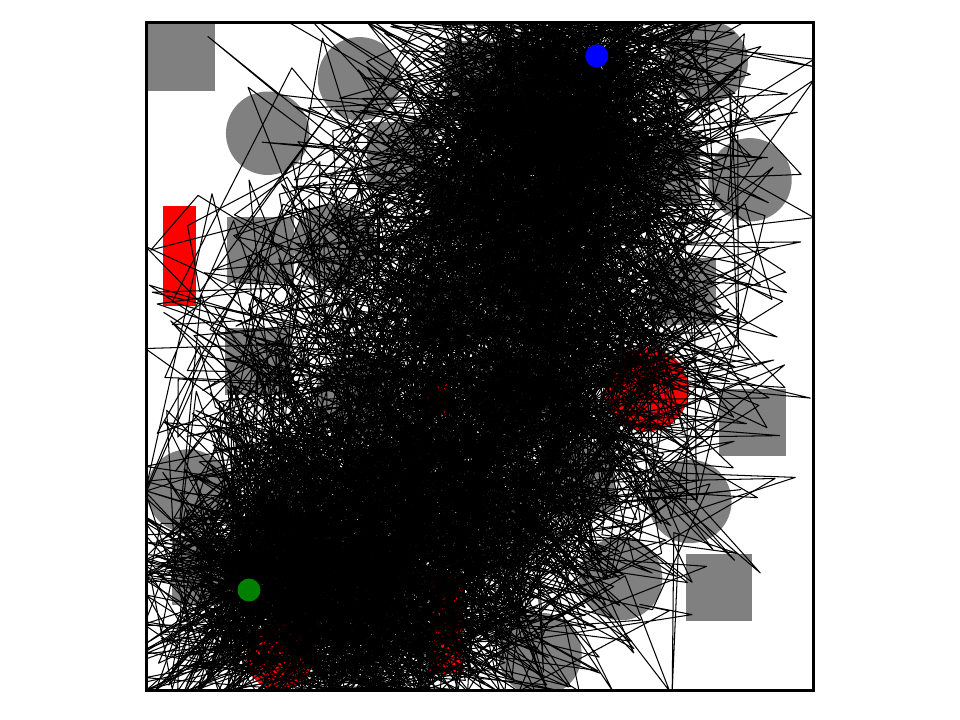}}
        \label{fig:diffusion:10}
    \hfill
  \subfloat[Diffusion Step 20]{%
        \includegraphics[width=0.191\linewidth]{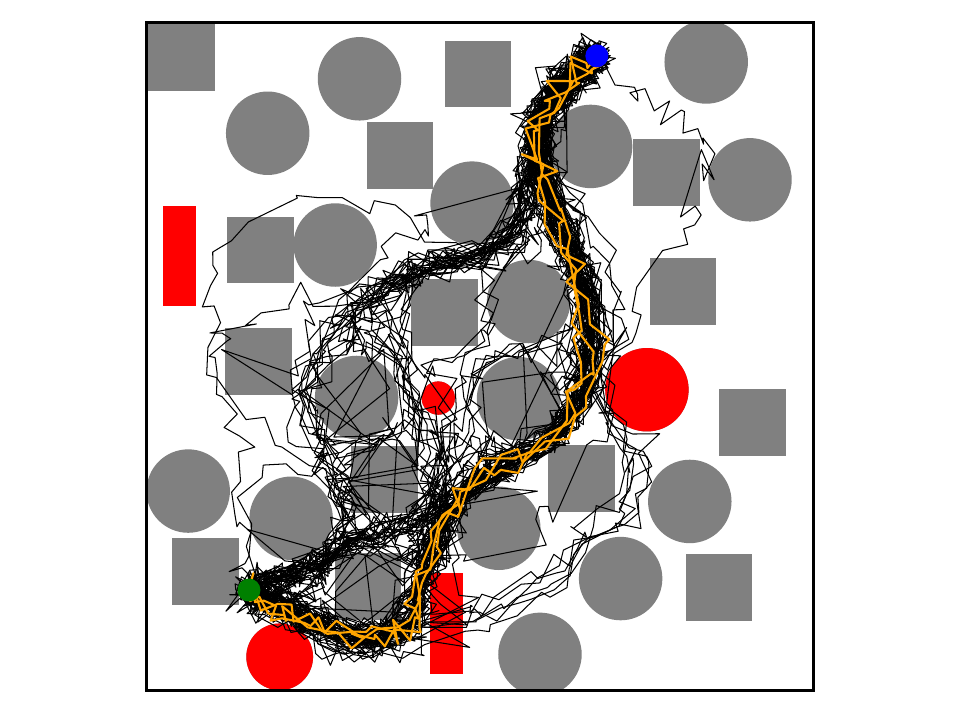}}
        \label{fig:diffusion:20}
    \hfill
  \subfloat[Diffusion Step 25]{%
        \includegraphics[width=0.191\linewidth]{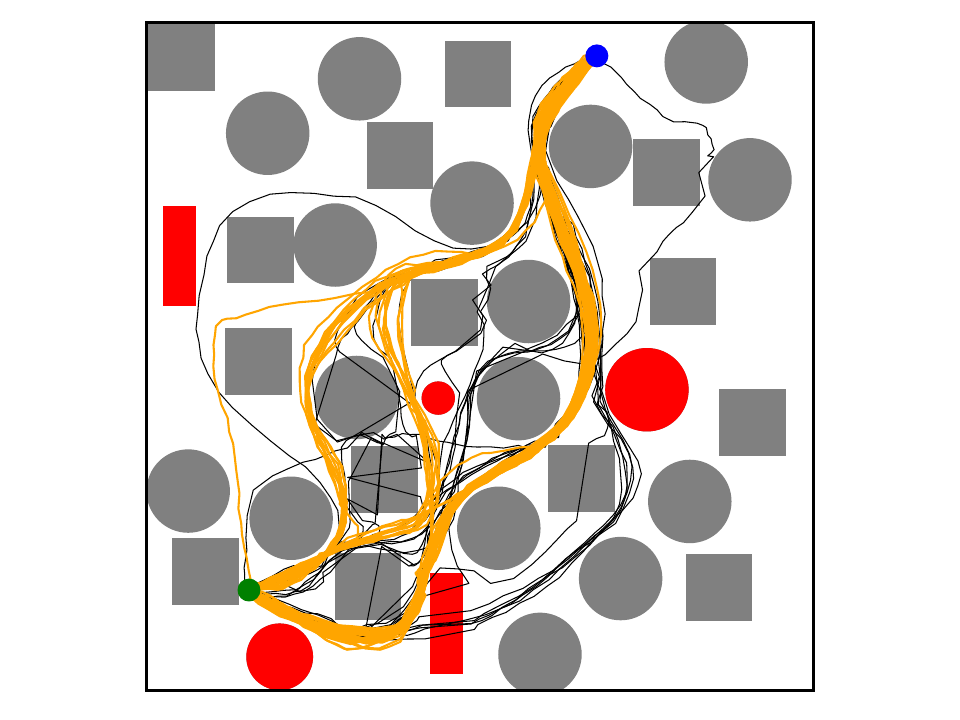}
        \label{fig:diffusion:25}
        }
  \vrule\
  \subfloat[CVAE]{%
       \includegraphics[width=0.191\linewidth]{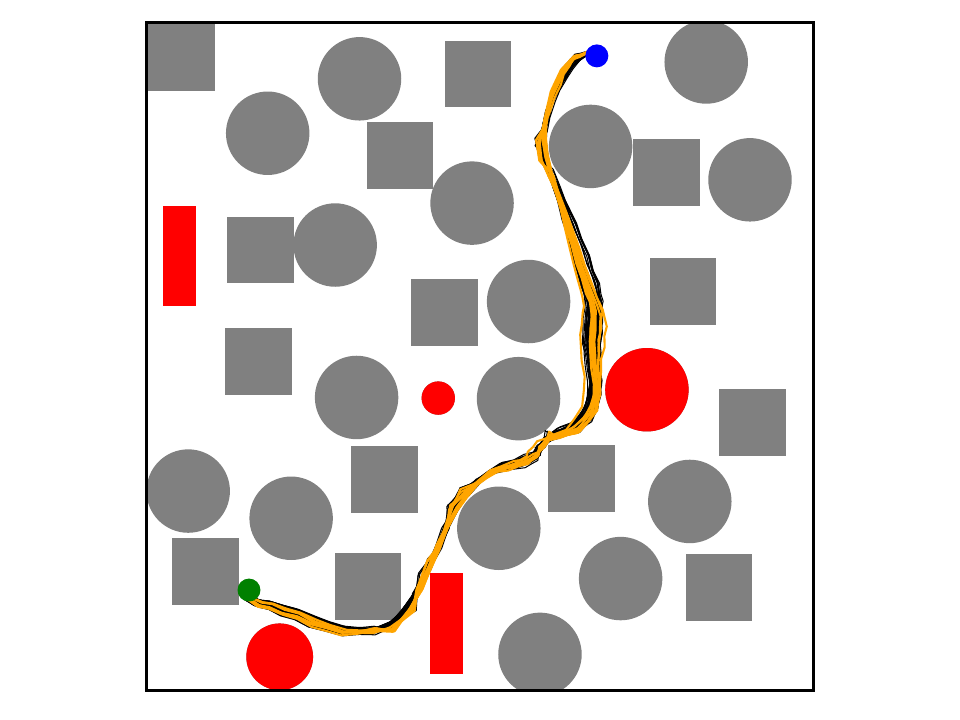}
       \label{fig:cvae-sampling}
       }
  \caption{
    (a)-(d) Diffusion steps of MPD on a batch of $100$ trajectories in the PointMass2D Dense - Extra Obstacles environment.
    Notice how noise transforms into multimodal, smooth and collision-free trajectories.
    (e) Trajectories generated by CVAEPosterior.
    Obstacles in red were not present in the training environment.
    Trajectories in orange are collision-free, black in collision.
    The start and goal configurations are in green and blue.
  }
  \label{fig:diffusion-sampling-pointmass2d}
\vspace{-0.15cm}
\end{figure*}

To verify the advantages of our approach, we perform experiments to answer the following questions: 
(1) Can the diffusion model learn collision-free high-dimensional and highly-multimodal trajectory distributions?
(2) How does the diffusion model prior compares to a commonly used Conditional Variational AutoEncoder (CVAE) as a trajectory generative model?
(3) Can our approach generate collision-free trajectories in the presence of obstacles not seen during training?
(4) Does merging sampling from the prior while biasing samples with cost gradients improve results, compared to other generative models that first sample from the prior and then optimize the likelihood? 
(5) Is the diffusion model a good prior for optimization-based planning algorithms?

\vspace{-0.1cm}
\subsection{Experimental Setup}

\textbf{Environments.}
We consider different environments reflecting increasing difficulty, as depicted in Fig.~\ref{fig:environments}.
PointMass2D Dense is a $2$D planar navigation environment with randomly placed spheres and boxes that a point mass robot has to navigate.
PointMass3D Maze Boxes is a $3$D maze environment with boxes and narrow passages.
Panda Spheres has a $7$-dof Franka Emika Panda robot and uniformly random spheres in its workspace, which benchmarks planners in higher-dimensional and non-trivial collision-free configuration space manifold.
Panda Shelf has a $7$-dof Panda robot placed on a working bench and with a nearby shelf, which replicates realistic tasks as a proof-of-concept of our method. 
The trajectory generative models (diffusion and CVAE) are trained with expert data generated from these environments.
To test the generalization capabilities of the learning-based methods, all environments are extended with additional sphere and box obstacles that are not present during training - identified with the suffix ``Extra Obstacles''.

\textbf{Tasks.}
In all environments, the task is to find a collision-free and smooth trajectory starting and finishing at random configuration-space positions $\jointpositionvec_{\text{start}}$ and $\jointpositionvec_{\text{goal}}$, which do not result in a task-space collision.
In the real-world Panda Shelf environment, besides finding a collision-free path, we additionally add an extra cost to maintain the end-effector orientation constant along the trajectory. 
Using \eqref{eq:distance_sethree}, the trajectory cost is ${c_{\text{EE}}(\trajectoryvec) = \sum_{k=0}^{H-1} d_{\sethree}\left( (\mT(\vq[k]), \mT_g[k]) \right) }$,
where $\mT(\vq[k])$ is the end-effector pose at waypoint $k$ and $\mT_g[k]$ the desired one.
Since we only want to enforce the orientations, we only set the desired goal orientation and copy the desired end-effector position from the waypoint position, i.e., $\mT(\vq[k]) = [\mR_k, \vp_k], \,\mT_g[k] = [\mR_g, \vp_k]$.

\textbf{Algorithms and Baselines.}
We denote the diffusion model trained over collision-free trajectories as \textsc{DiffusionPrior}, and our proposed approach of sampling with cost guidance (sec.~\ref{subsec:optimal-sampling-guidance}) as MPD.
These methods are compared against a set of baselines to showcase different properties.
A simple baseline is a sample-based method RRTConnect~\cite{kuffner2000rrtconnect}.
We modify this algorithm in several ways to fully use the GPU parallelization capabilities:
at the start of optimization, we pre-compute a buffer of collision-free configurations (the computation time of this operation is negligible due to parallelization);
a sample is removed from the buffer when added to the search tree;
the buffer is refilled when there are no samples left;
and computing the nearest neighbor node is done in parallel using the GPU.
These improvements lead to very fast planning for RRTConnect.

On the choice of the optimization-based planner, we use GPMP (i.e., GPMP2 in~\cite{Mukadam2018-gpmp-ijrr}) without an informative prior - a constant-velocity straight line connecting the start and goal configurations.
On conditional trajectory generative model as a baseline, we learn a Conditional Variational AutoEncoder (CVAE)~\cite{sohn2015cvae, ichter2018learning}, denoted as CVAEPrior.
For a fair comparison, the encoder and decoder networks are the same as the encoding and decoding parts of the U-Net diffusion model.
The conditioning variable is the start and goal configurations. 
These are stacked with the learned trajectory representation of the encoder and then passed through a neural network to encode a Gaussian posterior in the latent space.
We ran a hyperparameter search on the latent space dimension, learning rate, and the KL regularization multiplier and chose the model with the best validation loss on the same dataset used to train the diffusion model.
We analyze sampling from this prior first and afterward optimize the trajectory cost likelihood - this method is named CVAEPosterior.
The number of optimization steps is equivalent to the one MPD uses. 
We use the same cost temperatures $\lambda_i$ for both MPD and CVAEPosterior.
Finally, the learned diffusion and CVAE models can be used as a prior for GPMP, as well as a simple RRTConnect prior, which can be a way to provide an initial trajectory for optimization-based motion planning~\cite{ratliff2009chomp}.
These are referred to as [Prior]-GPMP.
If RRTConnect is used as a prior, we smoothen that solution using B-splines before passing it to GPMP.\\
\indent \textbf{Metrics.}
A set of commonly used metrics for planning evaluation are used.
For all metrics, we report the mean and standard deviation of motion planning results of $100$ random contexts (start and goal configuration-free positions) when sampling $100$ trajectories per context.
Note that for algorithms that use RRTConnect, the trajectories for each context are sampled sequentially since GPU batch parallelization is not trivial for these methods.
To assess the speed of RRTConnect in generating a single collision-free path, the reported times can be divided by $100$, making the algorithm, on average, faster than one second.
Time (\textsc{T}) - is the planner's computational time to generate/optimize the required number of trajectories.
Success (\textsc{S}) - is the success rate, which is $1$ if at least one of the trajectories in the batch is collision-free, and $0$ otherwise.
Intensity (\textsc{I}) - is the percentage of the waypoints that are in collision, which assesses the ability to generate almost-collision-free trajectories.
Path Length (\textsc{PL}) - is the trajectory length.
Waypoint variance (\textsc{Var}) - is the sum (along the trajectory dimension) of the pairwise L$2$-distance variance between waypoints at corresponding time steps.
This metric measures how multimodal (spread) the generated trajectories are. \\
\indent \textbf{Dataset generation and training.}
To generate multimodal and collision-free expert data trajectories in each environment, we sample $500$ random start and goal context configurations and $20$ trajectories per context.
We use RRTConnect to get a rough initial solution, then smoothen it using a B-spline, and run many optimization steps of Stochastic-GPMP~\cite{urain_2022_learning_implicit_priors} to create collision-free and smooth trajectories, similar as~\cite{Bhardwaj2020diffgpmp}.
This process is costly but done offline once.
The data is split into training and validation datasets ($5\%$ of the data).
The input to the diffusion and CVAE models is a trajectory of dimension $H\times d$, with $H$ the horizon and $d$ the state-space dimension, e.g., we used $H=64$ and $d=14$ in the panda environments.
The horizon value is a hyperparameter, but $64$ was sufficient for the motion planning tasks we considered.
The diffusion models are trained for $25$ diffusion steps with exponential scheduling (we found it to work better than linear scheduling~\cite{ho2020denoisingdiffusion}).
The models are trained using early stopping by 
inspecting the validation loss. \\
\indent \textbf{Implementation.}
For maximal parallelization, all environments, algorithms, and costs were implemented in PyTorch~\cite{pytorch2019} utilizing the GPU.
The experiments were conducted on a machine with an AMD EPYC $7453$ $28$-Core Processor and NVIDIA GeForce RTX $3090$.

\vspace{-0.2cm}
\subsection{Results in Simulation}

The summarized results from planning in simulated environments are detailed in Table~\ref{tab:benchmark}. 
To answer our first and second questions, we look into the environments from the training set and observe that DiffusionPrior can generate more collision-free trajectories than the baseline CVAEPrior.
E.g., in the PointMass2D Dense environment, the success rate is $98\%$ vs. $46\%$.
Moreover, the diffusion model can produce more multimodal trajectories, as the variance column shows.
Empirically, we observed that across different planning problems, the CVAE models tended to generate fewer modes than the ones from the diffusion.

To assess questions 3 and 4, we look into the environments with the suffix ``Extra Obstacles'', where new obstacles not seen during training are placed randomly in the environment.
First, we observe that the success rate by sampling only from the priors decreases, which is expected since they were not trained in these environments.
Note that in the Panda Shelf environment, the success rate of DiffusionPrior does not decrease from $100\%$, but the collision intensity increases from $3.6\%$ to $5.9\%$.
Observing the success rates of MPD and comparing them to CVAEPosterior, we see that using the diffusion model in combination with the cost gradients yields better results in terms of success rate and multimodality (as the variance measures the spread of the distribution).
This phenomenon can be seen in Fig.~\ref{fig:diffusion-sampling-pointmass2d}, where we compare the optimization with the diffusion and CVAE models.
A common criticism of diffusion models is their sampling time.
However, for motion planning, the computation times of both methods for the Panda environments are comparable since most of the cost is spent in the gradient computation of the cost and not in the diffusion sampling.

Finally, we check if the diffusion samples act as a good prior for GPMP.
Across all environments, using the model samples as initialization for GPMP generated higher success rates and multimodal trajectories compared to running GPMP with a constant velocity straight-line mean prior.
The low success rate of GPMP is more noticeable in environments where the collision-free space manifold is very complex, as in the PointMass2D Dense environment.
The larger compute times of algorithms using GPMP are due to running this algorithm with a large batch of trajectories since at every time-step it solves a trust region gradient.

The sampling times for RRTConnect include sequentially sampling trajectories using the GPU since, as noted before, parallelization of RRT-like algorithms for a batch is not trivial.
Even though RRTConnect also is a fast prior for GPMP, it generally produces high-jerk trajectories with higher path lengths, which would need more optimization steps of GPMP to achieve the same level of the DiffusionPrior.
\vspace{-0.2cm}
\subsection{Results in the Real World Panda Shelf}

We test our method in a real-world task where a robot moves a bottle while avoiding obstacles not present in the training set and maintaining a constant end-effector orientation.
We approximate the new obstacles (cf. Fig.~\ref{fig:real_robot_task}) as box models.
With the robot in gravity compensation mode, we record $3$ different initial configurations and one desired goal configuration.
For each start-goal pair, we sampled $10$ trajectories with the MPD, filtered the ones in collision, obtaining $25$ collision-free trajectories, and without selecting them with respect to any criteria, ran them in the real robot using a joint impedance controller.
We obtained $19/25$ collision-free trajectories, demonstrating our method's efficacy of generating diverse success solutions.
We hypothesize that the remaining $6$ trajectories that collided with the new obstacles are due to the approximated robot (spheres) collision model.

\section{Conclusion and Future Work}

In this paper, we proposed using diffusion models as priors for bootstrapping motion planning problems, via the planning-as-inference perspective. 
We parameterize a trajectory with waypoints and construct a generative model over the whole trajectory.
We train this model via supervised learning on motion plans generated with an optimal planner.
At inference time, instead of sampling trajectories from the prior to only initialize an optimization-based motion planner, we propose to use the guidance properties of diffusion models to concurrently sample from the prior and bias these samples towards regions of low cost (high likelihood).

Our results show several benefits of diffusion models.
Due to their modeling capabilities, diffusion models are better priors for motion planning because they can encode multimodal trajectories better than commonly used CVAEs.
Sampling from the learned prior while optimizing the likelihood leads to improved results in finding collision-free trajectories.
Future work will include continuing to extend the usage of diffusion models to encode different parametrizations of trajectories for robotic movements.

\bibliographystyle{biblio/IEEEtran}
\bibliography{biblio/IEEEabrv, main.bib}

\begin{thebibliography}{10}
\providecommand{\url}[1]{#1}
\csname url@rmstyle\endcsname
\providecommand{\newblock}{\relax}
\providecommand{\bibinfo}[2]{#2}
\providecommand\BIBentrySTDinterwordspacing{\spaceskip=0pt\relax}
\providecommand\BIBentryALTinterwordstretchfactor{4}
\providecommand\BIBentryALTinterwordspacing{\spaceskip=\fontdimen2\font plus
\BIBentryALTinterwordstretchfactor\fontdimen3\font minus
  \fontdimen4\font\relax}
\providecommand\BIBforeignlanguage[2]{{%
\expandafter\ifx\csname l@#1\endcsname\relax
\typeout{** WARNING: IEEEtran.bst: No hyphenation pattern has been}%
\typeout{** loaded for the language `#1'. Using the pattern for}%
\typeout{** the default language instead.}%
\else
\language=\csname l@#1\endcsname
\fi
#2}}

\bibitem{Lav2006planningalgorithms}
S.~M. LaValle, \emph{Planning Algorithms}.\hskip 1em plus 0.5em minus
  0.4em\relax Cambridge, U.K.: Cambridge University Press, 2006.

\bibitem{ratliff2009chomp}
N.~Ratliff, M.~Zucker, J.~A. Bagnell, and S.~Srinivasa, ``Chomp: Gradient
  optimization techniques for efficient motion planning,'' in \emph{IEEE
  International Conference on Robotics and Automation}, 2009.

\bibitem{Kalakrishnan_RAIIC_2011_stomp}
M.~Kalakrishnan, S.~Chitta, E.~Theodorou, P.~Pastor, and S.~Schaal, ``Stomp:
  Stochastic trajectory optimization for motion planning,'' in \emph{IEEE
  International Conference on Robotics and Automation}, 2011.

\bibitem{Elbanhawi2014sampling}
M.~Elbanhawi and M.~Simic, ``Sampling-based robot motion planning: A review,''
  \emph{IEEE Access}, vol.~2, pp. 56--77, 2014.

\bibitem{lynch2017modernrobotics}
K.~Lynch and F.~Park, \emph{\BIBforeignlanguage{English (US)}{Modern Robotics:
  Mechanics, Planning, and Control}}.\hskip 1em plus 0.5em minus 0.4em\relax
  Cambridge Univeristy Press, 2017.

\bibitem{Kavraki1996PRM}
L.~Kavraki, P.~Svestka, J.-C. Latombe, and M.~Overmars, ``Probabilistic
  roadmaps for path planning in high-dimensional configuration spaces,''
  \emph{IEEE Transactions on Robotics and Automation}, 1996.

\bibitem{Lavalle98rapidly-exploringrandom}
S.~M. Lavalle, ``Rapidly-exploring random trees: A new tool for path
  planning,'' 1998.

\bibitem{kuffner2000rrtconnect}
J.~Kuffner and S.~LaValle, ``Rrt-connect: An efficient approach to single-query
  path planning,'' in \emph{IEEE ICRA}, 2000.

\bibitem{karaman2011sampling}
S.~Karaman and E.~Frazzoli, ``Sampling-based algorithms for optimal motion
  planning,'' \emph{IJRR}, vol.~30, no.~7, pp. 846--894, 2011.

\bibitem{Dong2016-gpmp-rss}
J.~Dong, M.~Mukadam, F.~Dellaert, and B.~Boots, ``Motion planning as
  probabilistic inference using {G}aussian processes and factor graphs,'' in
  \emph{RSS}, 2016.

\bibitem{Mukadam2018-gpmp-ijrr}
M.~Mukadam, J.~Dong, X.~Yan, F.~Dellaert, and B.~Boots, ``Continuous-time
  gaussian process motion planning via probabilistic inference,'' \emph{Int. J.
  Robotics Res.}, vol.~37, no.~11, 2018.

\bibitem{hauser2010fastsmoothing}
K.~K. Hauser and V.~Ng-Thow-Hing, ``Fast smoothing of manipulator trajectories
  using optimal bounded-acceleration shortcuts.'' in \emph{ICRA}.\hskip 1em
  plus 0.5em minus 0.4em\relax IEEE, 2010, pp. 2493--2498.

\bibitem{urain_2022_learning_implicit_priors}
J.~Urain, A.~Le, A.~Lambert, G.~Chalvatzaki, B.~Boots, and J.~Peters,
  ``Learning implicit priors for motion optimization,'' in \emph{IEEE/RSJ
  International Conference on Intelligent Robots and Systems}, 2022.

\bibitem{wang2021survey}
J.~Wang \emph{et~al.}, ``A survey of learning-based robot motion planning,''
  \emph{IET Cyber-Systems and Robotics}, vol.~3, no.~4, pp. 302--314, 2021.

\bibitem{ichter2018learning}
B.~Ichter, J.~Harrison, and M.~Pavone, ``Learning sampling distributions for
  robot motion planning,'' in \emph{IEEE ICRA}, 2018.

\bibitem{Qureshi2018motionplanningnet}
A.~H. Qureshi, A.~Simeonov, M.~J. Bency, and M.~C. Yip, ``Motion planning
  networks,'' in \emph{IEEE ICRA}, 2019.

\bibitem{koert2016debato}
D.~Koert, G.~Maeda, R.~Lioutikov, G.~Neumann, and J.~Peters, ``Demonstration
  based trajectory optimization for generalizable robot motions,'' in
  \emph{IEEE-RAS Humanoids}, 2016, pp. 515--522.

\bibitem{rana2017towardsrobustskill}
M.~A. Rana, M.~Mukadam, S.~R. Ahmadzadeh, S.~Chernova, and B.~Boots, ``Towards
  robust skill generalization: Unifying learning from demonstration and motion
  planning,'' in \emph{CoRL}.\hskip 1em plus 0.5em minus 0.4em\relax {PMLR},
  2017.

\bibitem{le2021learning}
A.~T. Le \emph{et~al.}, ``Learning forceful manipulation skills from
  multi-modal human demonstrations,'' in \emph{IEEE/RSJ IROS}, 2021.

\bibitem{urain2022se3diffusion}
J.~Urain, N.~Funk, J.~Peters, and G.~Chalvatzaki, ``Se(3)-diffusionfields:
  Learning smooth cost functions for joint grasp and motion optimization
  through diffusion,'' in \emph{IEEE ICRA}, 2023.

\bibitem{ortiz2022structureddeep}
J.~Ortiz-Haro, J.-S. Ha, D.~Driess, and M.~Toussaint, ``Structured deep
  generative models for sampling on constraint manifolds in sequential
  manipulation,'' in \emph{CoRL}.\hskip 1em plus 0.5em minus 0.4em\relax PMLR,
  2022.

\bibitem{Takayuki2018imitationlearning}
T.~Osa, J.~Pajarinen, G.~Neumann, J.~A. Bagnell, P.~Abbeel, and J.~Peters, ``An
  algorithmic perspective on imitation learning,'' \emph{Found. Trends
  Robotics}, vol.~7, no. 1-2, pp. 1--179, 2018.

\bibitem{Sohl-Dickstein2015diffusion}
J.~Sohl{-}Dickstein, E.~A. Weiss, N.~Maheswaranathan, and S.~Ganguli, ``Deep
  unsupervised learning using nonequilibrium thermodynamics,'' in
  \emph{ICML}.\hskip 1em plus 0.5em minus 0.4em\relax JMLR.org, 2015.

\bibitem{ho2020denoisingdiffusion}
J.~Ho, A.~Jain, and P.~Abbeel, ``Denoising diffusion probabilistic models,'' in
  \emph{NeurIPS}.\hskip 1em plus 0.5em minus 0.4em\relax Curran Associates
  Inc., 2020.

\bibitem{song2019generativemodeling}
Y.~Song and S.~Ermon, ``Generative modeling by estimating gradients of the data
  distribution,'' in \emph{NeurIPS}, 2019.

\bibitem{rombach2021highresolution}
R.~Rombach, A.~Blattmann, D.~Lorenz, P.~Esser, and B.~Ommer, ``High-resolution
  image synthesis with latent diffusion models,'' 2021.

\bibitem{Kim_2022_CVPR}
G.~Kim, T.~Kwon, and J.~C. Ye, ``Diffusionclip: Text-guided diffusion models
  for robust image manipulation,'' in \emph{IEEE/CVF CVPR}, 2022.

\bibitem{ramesh2022dalle2}
\BIBentryALTinterwordspacing
A.~Ramesh \emph{et~al.}, ``Hierarchical text-conditional image generation with
  clip latents,'' 2022. [Online]. Available:
  \url{https://arxiv.org/abs/2204.06125}
\BIBentrySTDinterwordspacing

\bibitem{dhariwal2021diffusion}
P.~Dhariwal and A.~Nichol, ``Diffusion models beat gans on image synthesis,''
  2021.

\bibitem{johnson2021motionplanningtransformers}
\BIBentryALTinterwordspacing
J.~J. Johnson \emph{et~al.}, ``Motion planning transformers: A motion planning
  framework for mobile robots,'' 2021. [Online]. Available:
  \url{https://arxiv.org/abs/2106.02791}
\BIBentrySTDinterwordspacing

\bibitem{qureshi2018deepsmp}
A.~H. Qureshi and M.~C. Yip, ``Deeply informed neural sampling for robot motion
  planning,'' in \emph{{IEEE/RSJ} IROS}, 2018.

\bibitem{wang2020neuralrrt}
J.~Wang, W.~Chi, C.~Li, C.~Wang, and M.~Q.-H. Meng, ``Neural rrt*:
  Learning-based optimal path planning,'' \emph{IEEE T-ASE}, 2020.

\bibitem{grathwohl2020mcmc}
W.~Grathwohl, J.~Kelly, M.~Hashemi, M.~Norouzi, K.~Swersky, and D.~Duvenaud,
  ``No mcmc for me: Amortized sampling for fast and stable training of
  energy-based models,'' 2020.

\bibitem{Kapelyukh2022-dall-e-bot}
I.~Kapelyukh, V.~Vosylius, and E.~Johns, ``Dall-e-bot: Introducing web-scale
  diffusion models to robotics,'' 2022.

\bibitem{liu2022structdiffusion}
\BIBentryALTinterwordspacing
W.~Liu, T.~Hermans, S.~Chernova, and C.~Paxton, ``Structdiffusion:
  Object-centric diffusion for semantic rearrangement of novel objects,'' 2022.
  [Online]. Available: \url{https://arxiv.org/abs/2211.04604}
\BIBentrySTDinterwordspacing

\bibitem{findlay2022ddpmwalking}
\BIBentryALTinterwordspacing
E.~J.~C. Findlay, H.~Zhang, Z.~Chang, and H.~P.~H. Shum, ``Denoising diffusion
  probabilistic models for styled walking synthesis,'' 2022. [Online].
  Available: \url{https://arxiv.org/abs/2209.14828}
\BIBentrySTDinterwordspacing

\bibitem{carvalho2022conditionedsbm}
J.~Carvalho, M.~Baierl, J.~Urain, and J.~Peters, ``Conditioned score-based
  models for learning collision-free trajectory generation,'' in \emph{NeurIPS
  2022 Workshop on Score-Based Methods}, 2022.

\bibitem{janner2022diffuser}
M.~Janner, Y.~Du, J.~Tenenbaum, and S.~Levine, ``Planning with diffusion for
  flexible behavior synthesis,'' in \emph{ICML}, 2022.

\bibitem{ajay2022decision_diffuser}
\BIBentryALTinterwordspacing
A.~Ajay \emph{et~al.}, ``Is conditional generative modeling all you need for
  decision-making?'' 2022. [Online]. Available:
  \url{https://arxiv.org/abs/2211.15657}
\BIBentrySTDinterwordspacing

\bibitem{pmlr-vR4-attias03a}
H.~Attias, ``Planning by probabilistic inference,'' in \emph{Proceedings of the
  Ninth International Workshop on Artificial Intelligence and Statistics},
  vol.~R4.\hskip 1em plus 0.5em minus 0.4em\relax PMLR, 2003, pp. 9--16.

\bibitem{toussaint2009robusttrajopt}
M.~Toussaint, ``Robot trajectory optimization using approximate inference,'' in
  \emph{ICML}.\hskip 1em plus 0.5em minus 0.4em\relax Association for Computing
  Machinery, 2009.

\bibitem{levine2018rlcontrolasinference}
S.~Levine, ``Reinforcement learning and control as probabilistic inference:
  Tutorial and review,'' 2018.

\bibitem{peters2010reps}
J.~Peters, K.~M\"{u}lling, and Y.~Alt\"{u}n, ``Relative entropy policy
  search,'' in \emph{AAAI}.\hskip 1em plus 0.5em minus 0.4em\relax AAAI Press,
  2010.

\bibitem{watson2019i2c}
J.~Watson, H.~Abdulsamad, and J.~Peters, ``Stochastic optimal control as
  approximate input inference,'' in \emph{CoRL}.\hskip 1em plus 0.5em minus
  0.4em\relax {PMLR}, 2019.

\bibitem{watson22corl}
J.~Watson and J.~Peters, ``Inferring smooth control: Monte carlo posterior
  policy iteration with gaussian processes,'' in \emph{CoRL}, 2022.

\bibitem{Le__2022}
A.~T. Le, K.~Hansel, J.~Peters, and G.~Chalvatzaki, ``Hierarchical policy
  blending as optimal transport,'' in \emph{arXiv preprint arXiv:2212.01938},
  2022.

\bibitem{dong2018sparse}
J.~Dong, M.~Mukadam, B.~Boots, and F.~Dellaert, ``Sparse {G}aussian processes
  on matrix {L}ie groups: A unified framework for optimizing continuous-time
  trajectories,'' in \emph{IEEE ICRA}, 2018.

\bibitem{nichol2021improvedddpm}
A.~Q. Nichol and P.~Dhariwal, ``Improved denoising diffusion probabilistic
  models,'' in \emph{ICML}.\hskip 1em plus 0.5em minus 0.4em\relax {PMLR},
  2021.

\bibitem{Dhariwal2021diffusionbeatsgans}
P.~Dhariwal and A.~Q. Nichol, ``Diffusion models beat gans on image
  synthesis,'' in \emph{NeurIPS}, 2021, pp. 8780--8794.

\bibitem{pytorch2019}
A.~Paszke \emph{et~al.}, ``Pytorch: An imperative style, high-performance deep
  learning library,'' in \emph{NeurIPS}.\hskip 1em plus 0.5em minus 0.4em\relax
  Curran Associates, Inc., 2019.

\bibitem{sola2018micro}
J.~Sola, J.~Deray, and D.~Atchuthan, ``A micro lie theory for state estimation
  in robotics,'' \emph{arXiv preprint arXiv:1812.01537}, 2018.

\bibitem{barfoot2014batch}
T.~D. Barfoot, C.~H. Tong, and S.~S{\"a}rkk{\"a}, ``Batch continuous-time
  trajectory estimation as exactly sparse gaussian process regression.'' in
  \emph{RSS}, 2014.

\bibitem{sohn2015cvae}
K.~Sohn, H.~Lee, and X.~Yan, ``Learning structured output representation using
  deep conditional generative models,'' in \emph{NeurIPS}, vol.~28.\hskip 1em
  plus 0.5em minus 0.4em\relax Curran Associates, Inc., 2015.

\bibitem{Bhardwaj2020diffgpmp}
M.~Bhardwaj, B.~Boots, and M.~Mukadam, ``Differentiable gaussian process motion
  planning,'' in \emph{{IEEE} ICRA}, 2020.

\end{thebibliography}

\end{document}